\documentclass[journal]{IEEEtran}

\usepackage{cite}

\usepackage[pdftex]{graphicx}
\DeclareGraphicsExtensions{.pdf,.jpeg,.png,.jpg}

\usepackage{amssymb}
\usepackage{pifont}
\usepackage{amsmath}
\usepackage[linesnumbered,ruled]{algorithm2e}

\usepackage{array}

\usepackage[caption=false,font=footnotesize]{subfig}

\usepackage{dblfloatfix}

\ifCLASSOPTIONcaptionsoff
  \usepackage[nomarkers]{endfloat}
  \let\MYoriglatexcaption\caption
  \renewcommand{\caption}[2][\relax]{\MYoriglatexcaption[#2]{#2}}
\fi

\usepackage{url}

\usepackage{multirow}
\usepackage{xcolor}
\usepackage{xspace}
\makeatletter
\DeclareRobustCommand\onedot{\futurelet\@let@token\@onedot}
\def\@onedot{\ifx\@let@token.\else.\null\fi\xspace}

\def\etal{\emph{et al}\onedot}
\makeatother

\newcommand{\xmark}{\ding{55}}

\begin{document}

\title{Fast End-to-End Trainable Guided Filter}

\author{Huikai~Wu,~\IEEEmembership{Student Member,~IEEE,}
		Shuai~Zheng,~Junge~Zhang,~\IEEEmembership{Member,~IEEE,}\\
        and~Kaiqi~Huang,~\IEEEmembership{Senior Member,~IEEE}%
\thanks{H. Wu and J. Zhang are with Institute of Automation, Chinese Academy of Sciences, Beijing 100190, China, and also with the University of Chinese Academy of Sciences, Beijing 100049, China. E-mail: huikai.wu@nlpr.ia.ac.cn, jgzhang@nlpr.ia.ac.cn}%
\thanks{S. Zheng is with eBay Research. The work was conducted while the author was at the University of Oxford. E-mail: shuzheng@ebay.com}%
\thanks{K. Huang is with Institute of Automation, Chinese Academy of Sciences, Beijing 100190, China, and also with the University of Chinese Academy of Sciences, Beijing 100049, China, and the CAS Center for Excellence in Brain Science and Intelligence Technology, 100190. E-mail: kqhuang@nlpr.ia.ac.cn}%
\thanks{A preliminary version of this work~\cite{Wu_2018_CVPR} appeared in IEEE Conference on Computer Vision and Pattern Recognition, 2018. The code is available at https://github.com/wuhuikai/DeepGuidedFilter.}
\thanks{This work is funded by the National Key Research and Development Program of China (Grant 2016YFB1001004 and Grant 2016YFB1001005), the National Natural Science Foundation of China (Grant 61673375, Grant 61721004 and Grant 61403383) and the Projects of Chinese Academy of Sciences (Grant QYZDB-SSW-JSC006 and Grant 173211KYS-B20160008). The authors would like to thank Patrick P{\'e}rez and Philip Torr for their helpful suggestions.}}

\markboth{}%
{}

\maketitle

\begin{abstract}
Dense pixel-wise image prediction has been advanced by harnessing the capabilities of Fully Convolutional Networks (FCNs).
One central issue of FCNs is the limited capacity to handle joint upsampling.
To address the problem, we present a novel building block for FCNs, namely guided filtering layer, which is designed for efficiently generating a high-resolution output given the corresponding low-resolution one and a high-resolution guidance map.
Such a layer contains learnable parameters, which can be integrated with FCNs and jointly optimized through end-to-end training.
To further take advantage of end-to-end training, we plug in a trainable transformation function for generating the task-specific guidance map.
Based on the proposed layer, we present a general framework for pixel-wise image prediction, named deep guided filtering network (DGF).
The proposed network is evaluated on five image processing tasks. 
Experiments on MIT-Adobe FiveK Dataset demonstrate that DGF runs 10-100 times faster and achieves the state-of-the-art performance.
We also show that DGF helps to improve the performance of multiple computer vision tasks.
\end{abstract}


\IEEEpeerreviewmaketitle

\section{Introduction}
\begin{figure}[!t]
	\centering
	
	\includegraphics[height=0.28\linewidth]{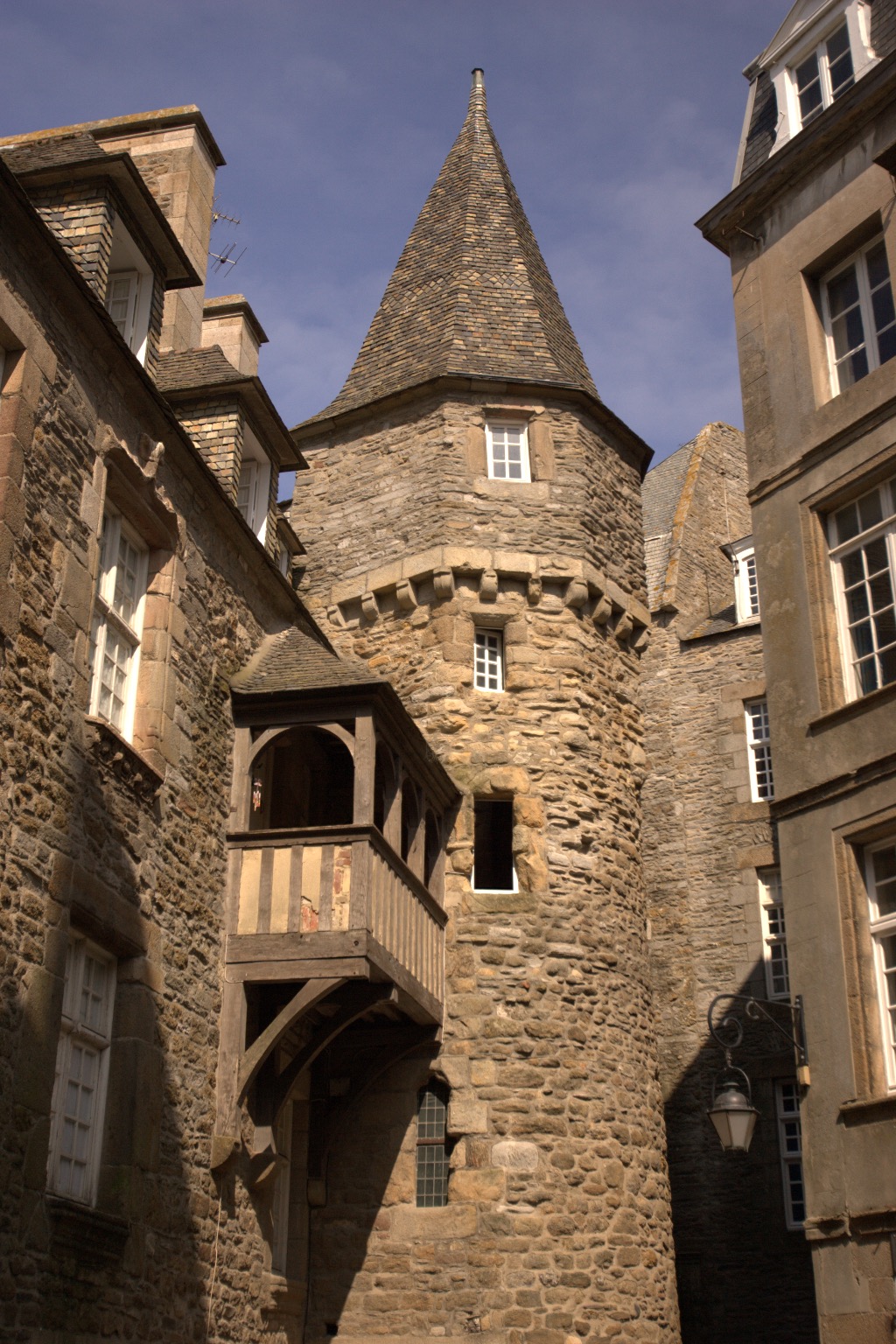}\hfill
	\includegraphics[height=0.28\linewidth]{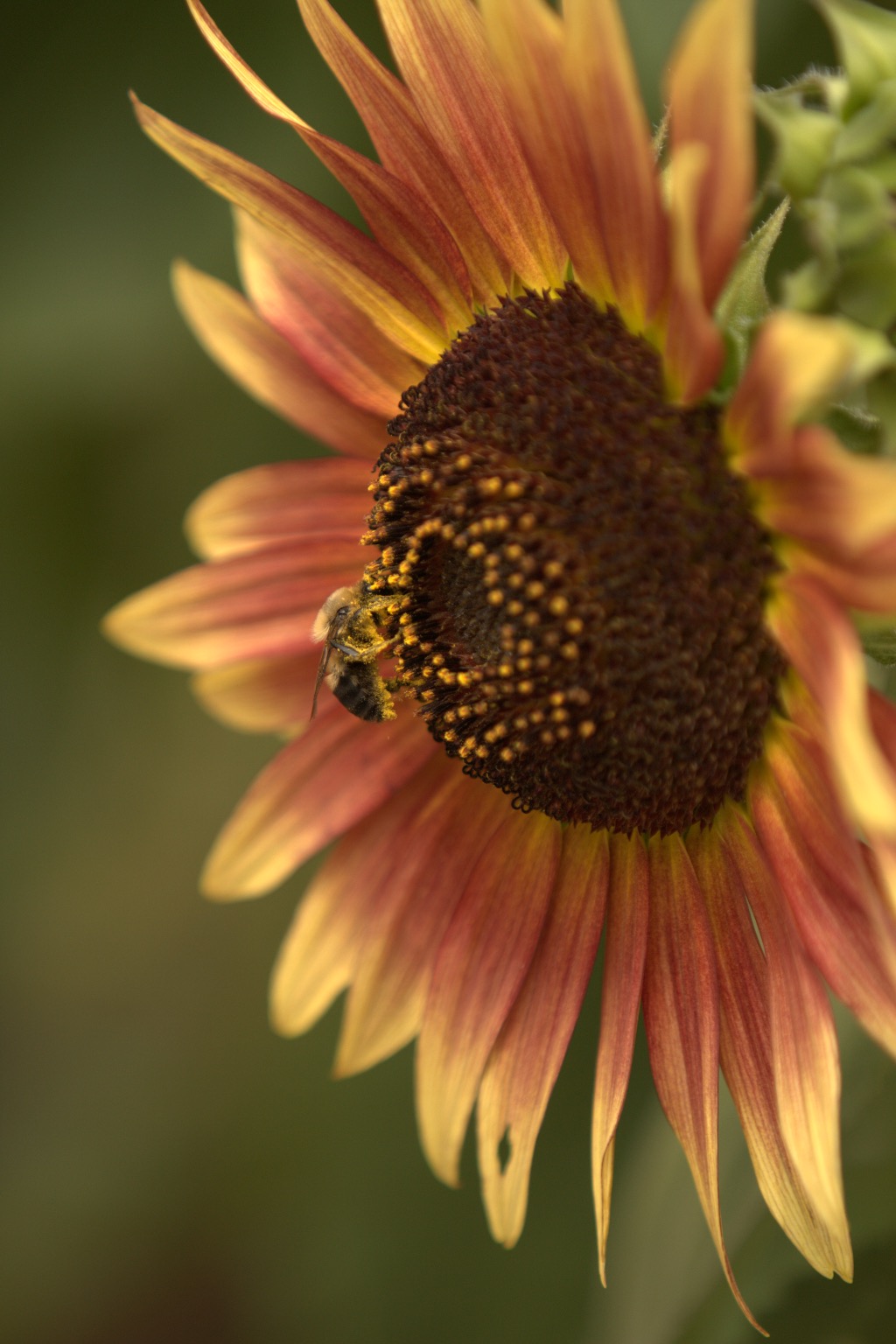}\hfill
	\includegraphics[height=0.28\linewidth]{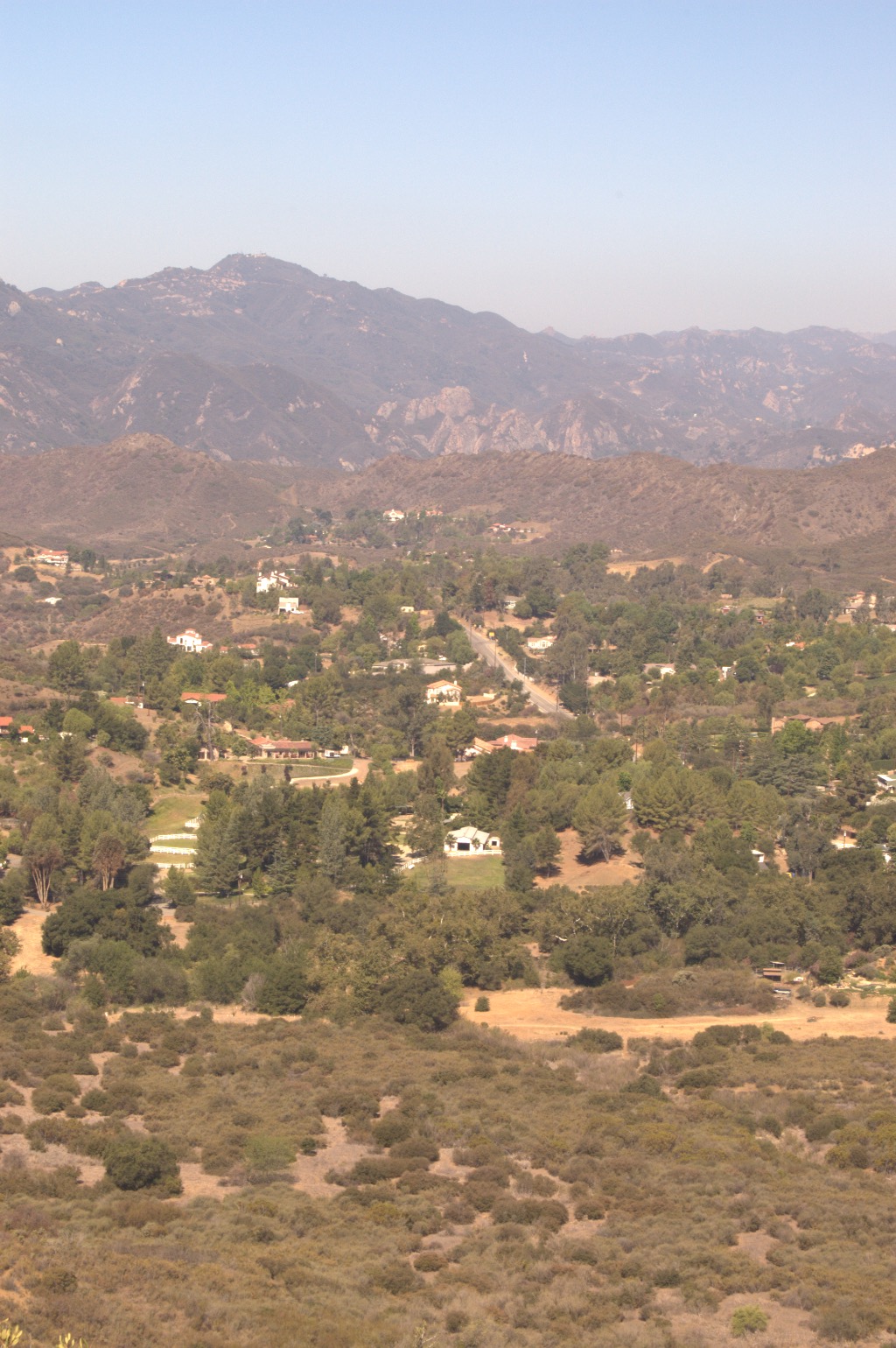}\hfill
	\includegraphics[height=0.28\linewidth]{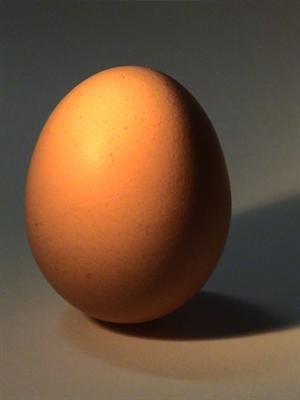}\hfill
	\includegraphics[height=0.28\linewidth]{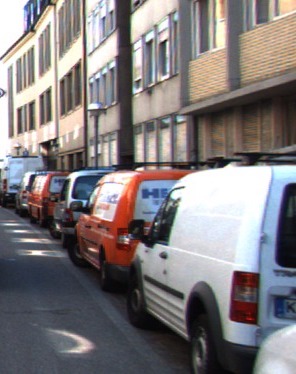}
	\\
	\vspace{0.1em}
	\includegraphics[height=0.28\linewidth]{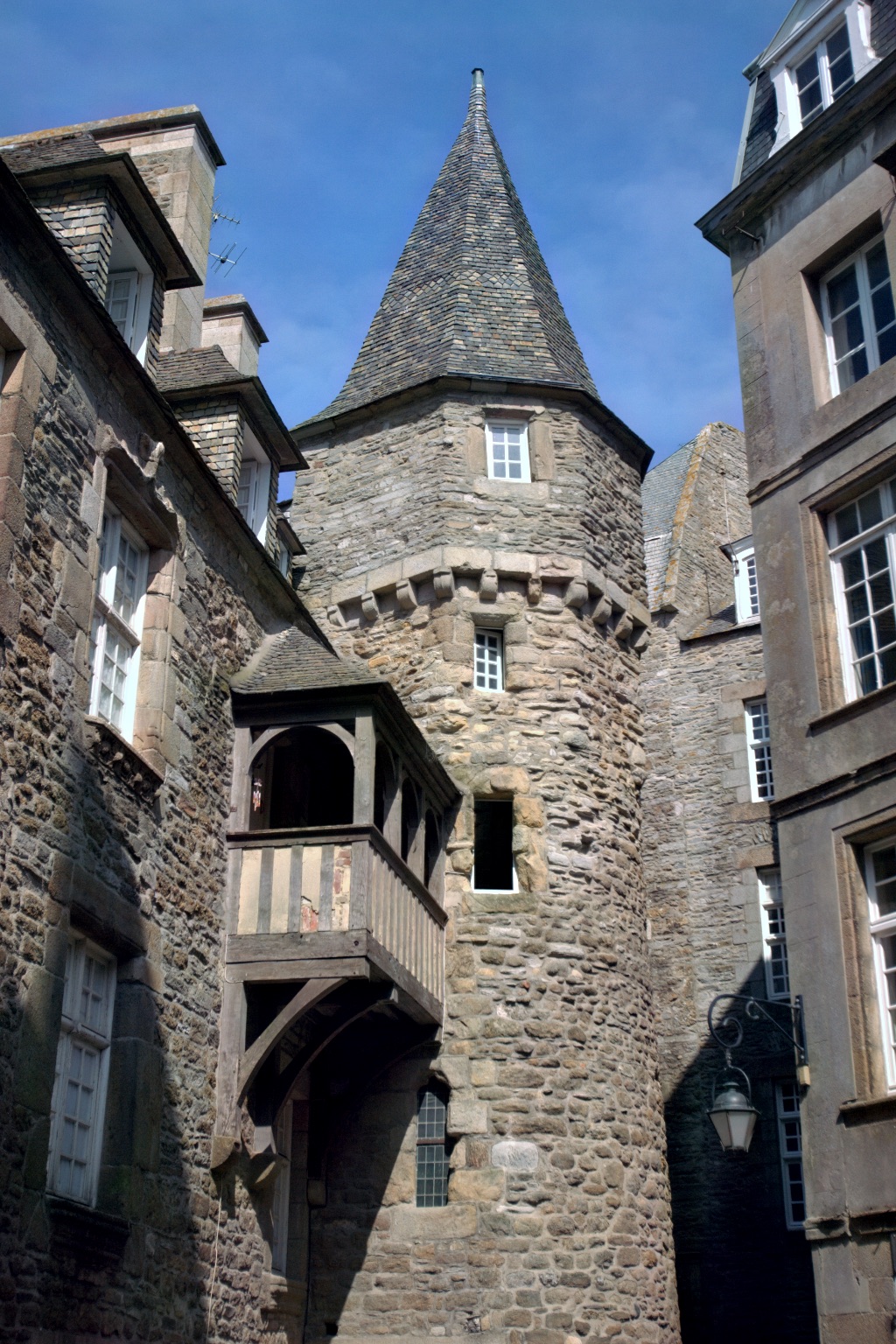}\hfill
	\includegraphics[height=0.28\linewidth]{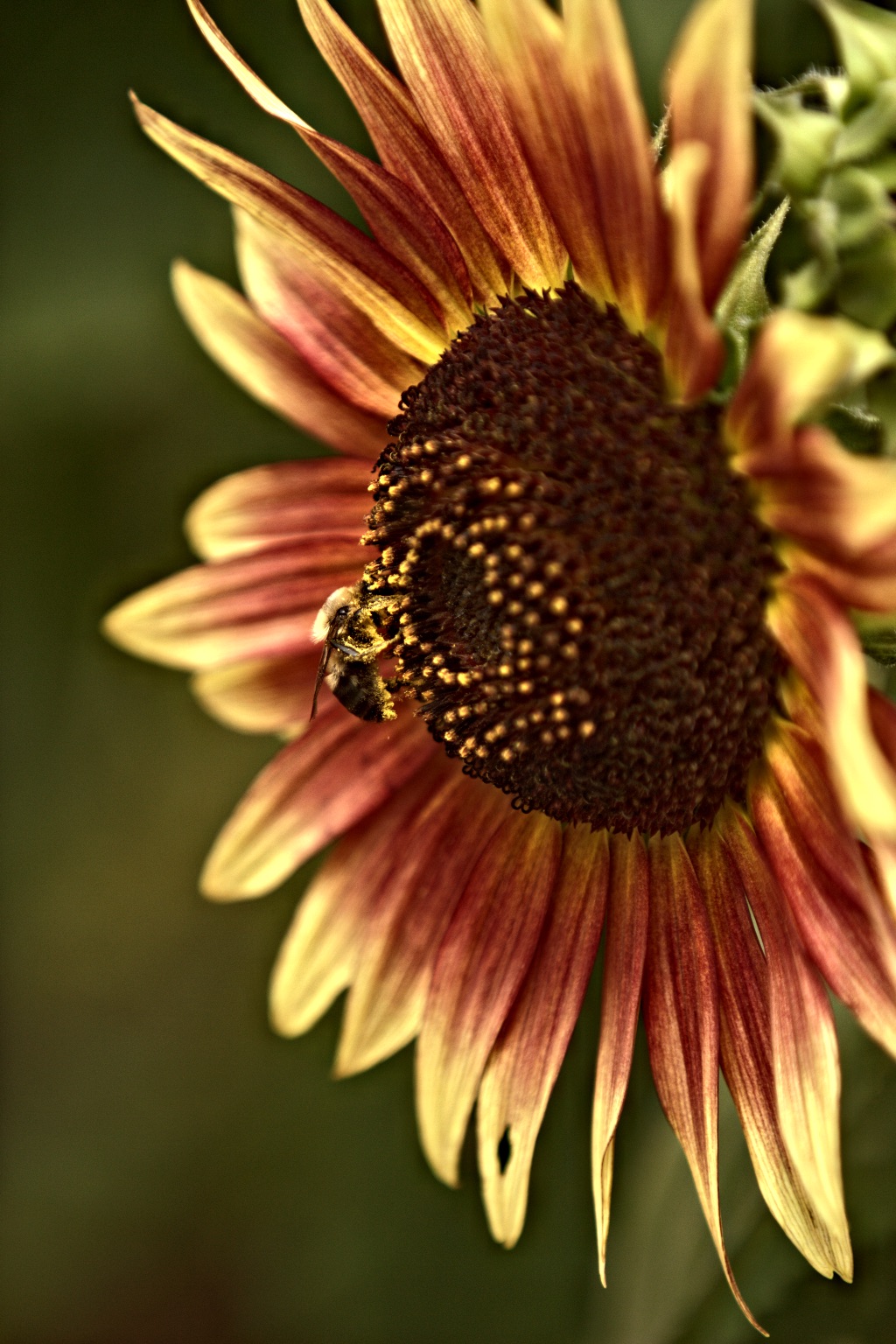}\hfill
	\includegraphics[height=0.28\linewidth]{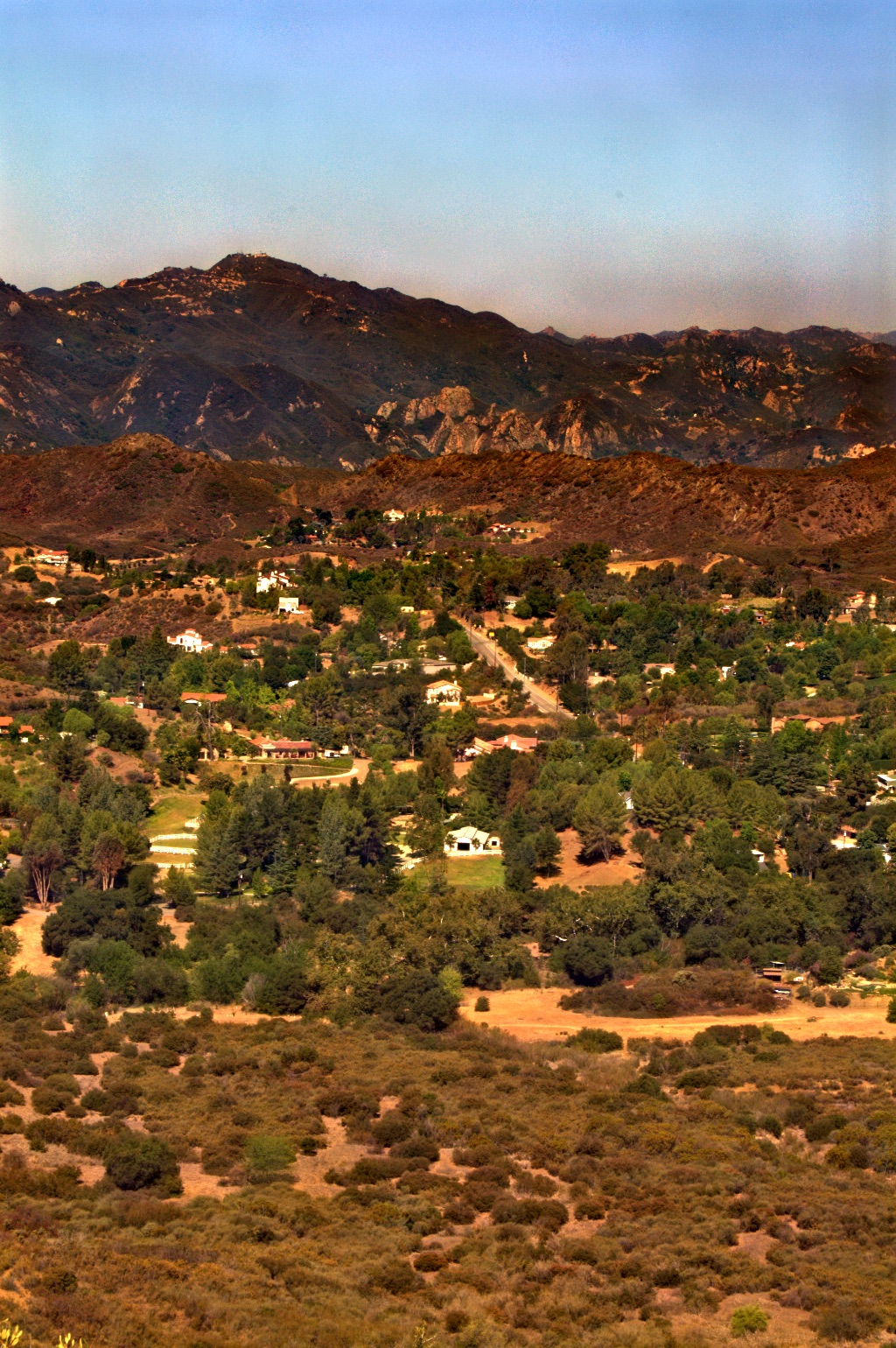}\hfill	
	\includegraphics[height=0.28\linewidth]{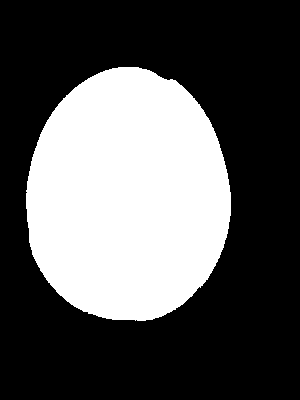}\hfill
	\includegraphics[height=0.28\linewidth]{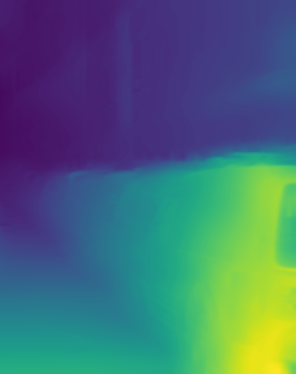}
	
	\caption{\textbf{Example Results of Deep Guided Filtering Network}. The top row shows the input images, and the bottom row presents the corresponding outputs. From left to right: image retouching~\cite{fivek}, multi-scale detail manipulation~\cite{Farbman/tog2008}, non-local dehazing~\cite{Berman/cvpr2016}, saliency detection~\cite{hou2017deeply}, and depth estimation~\cite{godard2017unsupervised}. Best viewed in color.}
	\label{fig:vis}
\end{figure}
\IEEEPARstart{D}{ense} pixel-wise image prediction is a fundamental image processing and computer vision problem and has a wide range of applications. In image processing, dense pixel-wise image prediction enables smoothing an image while preserving the edges~\cite{Xu/siggraph2011,Xu/siggraph2012,Zhang/eccv2014}, enhancing the details of an image~\cite{Farbman/tog2008,Subr/siggraph2009}, transferring the style from a reference image~\cite{Aubry/siggraph2014,Bae/siggraph2006}, dehazing the photos~\cite{Berman/cvpr2016,He/pami2011,Fattal/siggraph2008,Fattal/siggraph2014}, and retouching the images for global tonal adjustment~\cite{fivek}. In computer vision, pixel-wise image prediction not only addresses the problem of segmenting an image into semantic parts~\cite{He/cvpr2004,Shotton/eccv2006,cheng2017locality}, but also helps to estimate depth from a single image~\cite{Saxena/nips2005}, and detect the most salient object in an image~\cite{Liu/cvpr2007,Li/cvpr2015}. 

Recent methods~\cite{gharbi2017deep,Chen/iccv2017,chen2016deeplab} usually employ Fully Convolutional Networks (FCNs) for these applications, achieving state-of-the-art performance.
However, FCNs usually have a huge computational complexity and memory usage on high-resolution input images, which limits the deployment of pixel-wise image prediction algorithms in real-world applications.
To accelerate FCNs, we present a general framework by following a coarse-to-fine fashion, which firstly downsamples the input image, executes the algorithm at low resolution, and then upsamples the result back to the original resolution.
The main challenge is restoring the low-resolution output to the original resolution with rich details and sharp edges.

This challenge can be formulated as joint upsampling, which aims at generating a high-resolution output given the corresponding low-resolution one and a high-resolution guidance map.
However, existing building blocks of FCNs have limited capability to handle such a problem.
To enhance the ability of FCNs for joint upsampling, we propose to reformulate the widely used guided filter~\cite{He/2013} into a fully differentiable building block, which can be (1) jointly trained with FCNs, (2) adapted for different tasks by learnable parameters, and (3) directly supervised by high-resolution ground truth.

To this end, we propose a novel building block for FCNs named guided filtering layer.
Concretely, the original guided filter is expressed as a computational graph consisting of dilated convolutions and pointwise convolutions with learnable parameters, which can adaptively evolve for different tasks.
A trainable transformation function is introduced into the proposed layer, which can generate a task-specific guidance map.
As a result, all the parameters of a guided filtering layer can be learned in a data-driven manner through end-to-end training.
Moreover, such a layer can be easily integrated with a pre-defined FCN without extra efforts.
By equipping FCNs with guided filtering layer, we present a general framework for pixel-wise image prediction tasks named Deep Guided Filtering Network (DGF), which can largely reduce the computational complexity and memory usage.
The proposed framework can be widely employed for many image processing and computer vision tasks, as shown in Fig~\ref{fig:vis}.
Experiments show that DGF achieves the state-of-the-art performance in quality, speed, and memory usage.

In summary, the main contribution of this paper is that (1) we develop an end-to-end trainable guided filtering layer with learnable parameters and a trainable guidance map, which enhances the ability of FCNs for joint upsampling; (2) by combining with FCNs, the proposed layer significantly improves the state-of-the-art results in multiple image processing tasks, and runs $10\text{-}100\times$ faster than the alternatives; and (3) additional experiments show that our approach generalizes well to many computer vision tasks and achieves significant improvements over baseline methods.

An early version of this paper~\cite{Wu_2018_CVPR} appeared in IEEE Conference on Computer Vision and Pattern Recognition (2018), to which we have made substantial extensions.
The improvements are shown below:
(1) \cite{Wu_2018_CVPR} formulate the original guided filter into a series of spatially varying linear transformation matrices without any learnable parameters.
In this paper, we reformulate the original guided filter into a block of dilated convolutions and pointwise convolutions with learnable parameters.
Such a formulation enables guided filtering layer to fit a specific task through end-to-end training.
(2) Based on the improved guided filtering layer, we further boosted the performance of DGF in five image processing tasks. 
(3) We conduct a systematic ablation study on five image processing tasks to analysis the influence of each hyper-parameter in DGF. 
(4) We demonstrate the upper bound of the proposed layer's ability in joint upsampling through a comprehensive experiment.
(5) Both the training code and testing code are released for reproducing the experimental results in this paper and supporting further research as well as other applications.

\section{Related Work}
\subsection{Joint Upsampling}
The most related works to our method are along the direction of joint upsampling.
Many algorithms have been developed to tackle this problem.
Joint bilateral upsampling~\cite{Kopf/siggraph2007} applies a bilateral filter~\cite{Tomasi/iccv1998} to the high-resolution guidance map, resulting in a piecewise-smoothing high-resolution output.
The underlying bilateral filter usually requires a large amount of computation resources.
Thus, many methods~\cite{Adams/siggraph2009,Adams/eg2009,Gastal/siggraph2012} are presented to reduce the computation complexity.
Build on joint bilateral upsampling, Barron~\etal~\cite{barron2016fast} present a new form of bilateral-space optimization that efficiently solves a regularized least-squares optimization problem to produce an output that is bilateral-smooth and close to the input.
Gharbi~\etal~\cite{gharbi2015transform} first compute a description of the transformation from a highly compressed input to output. Then a high-fidelity approximation of the output can be constructed by applying the recipe to the high-quality input.
Similarly, bilateral guided upsampling~\cite{chen2016bilateral} fits an image operator with a grid of local affine models on the low-resolution input/output pair firstly.
The high-resolution output is then generated by applying the local affine model to the high-resolution input image.
This method serves as a post-processing operation, while our approach can be jointly trained with the entire FCN.
Deep bilateral learning~\cite{gharbi2017deep} integrate bilateral filter with FCNs, which can be jointly learned through end-to-end training.
However, this method requires producing affine coefficients before obtaining outputs, which lacks direct supervision from the ground truth.
For computer vision tasks, the number of affine coefficients is usually very large, which becomes the bottleneck of performance and speed.
Besides bilateral filter, guided filter~\cite{He/2013} is also widely used in joint upsampling, which derived from a local linear model and computes the filtering output by considering the content of a guidance image.
Compared to it, our method is formulated as a fully differentiable building block with learnable parameters, which can be jointly trained with FCNs and adaptively adjusted according to a specific task.
Similarly, Yuan~\etal~\cite{yuan2011high} employ a locally-affine model to relate patches from low-resolution RAW images to high-resolution JPEG images.

The above methods are based on edge-preserving local filters.
Differently, other methods~\cite{rudin1992nonlinear,farbman2008edge,yan2013cross} produce high-resolution outputs by optimizing manually designed objective functions involving all or many pixels.
The objective functions typically consist of data terms and regularization terms like total variation (TV)~\cite{rudin1992nonlinear}, weighted least squares (WLS)~\cite{farbman2008edge}, and scale map scheme~\cite{yan2013cross}.
Following these methods, Shen~\etal~\cite{shen2015mutual} propose mutual-structure to reserve the structural information that is contained in both images.
Similarly, Ham~\etal~\cite{ham2015robust} formulate the issue as a non-convex optimization problem, which is solved by the majorization-minimization algorithm.
Compared to our method, the main drawbacks of these methods are (1) they rely on hand-designed objective functions, and (2) they are usually time-consuming. 

\subsection{Deep Learning based Image Filter}
Recently, deep learning based methods are proposed in image processing tasks, which largely advanced the state-of-the-art performance.
Such tasks include image denoising~\cite{burger2012image}, image demosaicking~\cite{gharbi2016deep}, image deblurring~\cite{xu2014deep}, image matting~\cite{shen2016deep}, rain drop removal~\cite{eigen2013restoring}, image dehazing~\cite{ren2016single}, and image colorization~\cite{iizuka2016let}.

The above methods mainly focus on solving one specific image processing task.
Differently, some other works~\cite{xu2015deep,liu2016learning,isola2017image} aim at approximating a general class of operators. 
Xu~\etal~\cite{xu2015deep} employ deep neural networks to approximate a variety of edge-preserving filters with a gradient-domain training procedure, while Liu~\etal~\cite{liu2016learning} combine a convolutional network and a set of recurrent networks to approximate various image filters.

Xu~\etal~\cite{xu2015deep} and Liu~\etal~\cite{liu2016learning} deploy neural networks to generate high-resolution output directly, accelerating the operation by dedicatedly designed network architectures.
Similarly, Chen~\etal~\cite{Chen/iccv2017} propose context aggregation networks to accelerate a wide variety of image processing operators, which performs superior to the prior works~\cite{xu2015deep,liu2016learning,chen2016bilateral,johnson2016perceptual,Isola/cvpr2017}, achieving the best results regarding speed and accuracy.
Our approach is complementary to this method, which can deliver comparable or better results and runs $10\text{-}100\times$ faster.

Compared to all the related works, the proposed guided filtering layer can be end-to-end trained with the entire network and generalize well across different tasks ranging from image processing to computer vision, while achieving the state-of-the-art performance in both quality and speed.

\section{Guided Filtering Layer}
\subsection{Problem Formulation}
Given a high-resolution image $I_h$ and the corresponding low-resolution output $O_l$, joint upsampling aims at generating a high-resolution output $O_h$ that is visually similar to $O_l$ and preserves the edges and details from $I_h$.
In the literature of joint upsampling, guided filter~\cite{He/2013} is one of the most widely used algorithms that has shown better performance regarding the trade-off between speed and accuracy.

\subsection{Guided Filter Revisited}
\label{section:dgf_s}
To address joint upsampling, guided filter~\cite{He/2013} takes a low-resolution image $I_l$, the corresponding high-resolution image $I_h$, and a low-resolution output $O_l$ as inputs, producing the high-resolution output $O_h$.
Concretely, $A_l$ and $b_l$ are firstly obtained by minimizing a reconstruction error between $\hat{O}_l$ and $O_l$, where $\hat{O}_l$ subjects to a local linear model:
\begin{equation}\label{eq:local_linear_model}
    \hat{O}_l^i = A_l^k I_l^i + b_l^k, \forall i \in \omega_k.
\end{equation}
$\omega_k$ is the $k$-th local square window on $I_l$, and $I_l^i$ is the $i$-th pixel inside $\omega_k$.
$A_h$ and $b_h$ are then produced by upsampling $A_l$ and $b_l$.
The high-resolution output $O_h$ is finally generated by a linear transformation model:
\begin{equation}
    O_h = A_h * I_h + b_h,
\end{equation}
where $*$ is element-wise multiplication.

\begin{figure}[!t]
	\centering
	\includegraphics[width=\linewidth]{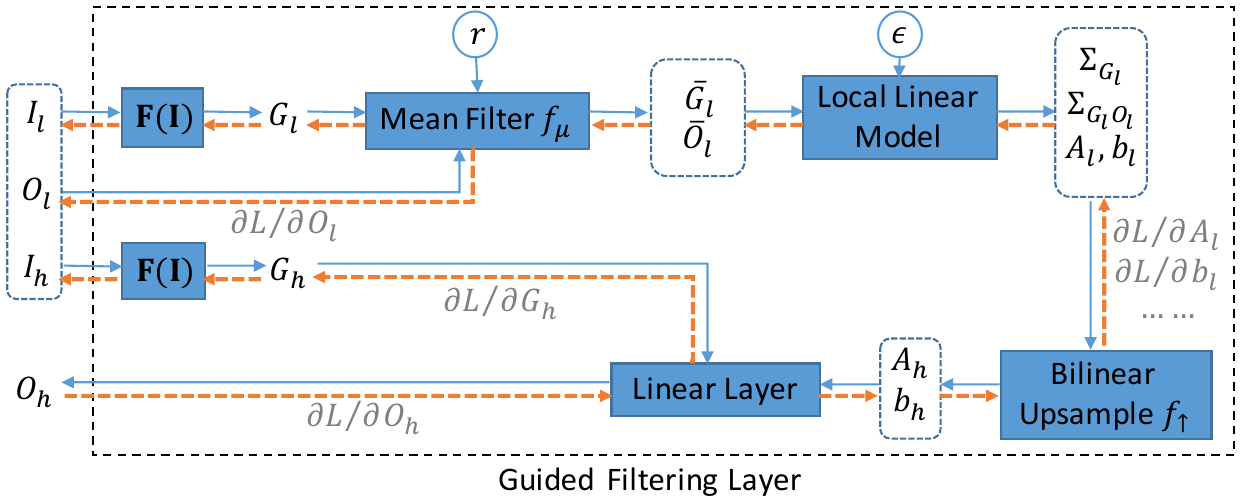}

	\caption{\textbf{Computation Graph of Guided Filtering Layer.} Guided filtering layer takes low-resolution image $I_l$, high-resolution image $I_h$ and low-resolution output $O_l$ as inputs, generating the high-resolution output $O_h$. Compared to guided filter~\cite{He/2013}, the proposed layer is reformulated into a fully differentiable block and employs $F(I)$ to generate task-specific guidance map.}
	\label{fig:guided_filter}
\end{figure}

\subsection{Fully Differentiable Guided Filter}
\label{section:dgf_b}
The original guided filter can only be employed as a post-processing operation, which is not differentiable and cannot be end-to-end trained with FCNs.
To enhance the ability of FCNs for joint upsampling, we propose a novel building block by reformulating guided filter into a fully differentiable layer.
Such a layer, named guided filtering layer, can be jointly trained with FCNs from scratch, and directly supervised by high-resolution targets. 

The computation graph of guided filtering layer is shown in Figure~\ref{fig:guided_filter}.
$A_l$ and $b_l$ are obtained by employing mean filter $f_{\mu}$ and local linear model to $I_l$ and $O_l$, where $f_{\mu}$ is implemented as a box filter to reduce the computation complexity.
$A_h$ and $b_h$ are then generated by bilinear upsampling $f_\uparrow$.
$O_h$ is finally produced by a linear layer taking $A_h$, $b_h$ and $I_h$ as inputs.
$r$ is the radius of $f_{\mu}$ and $\epsilon$ is the regularization term, which are set to be $1$ and $1e\text{-}8$ by default.

The equations for propagating the gradients through guided filtering layer are shown in Algorithm~\ref{algo:backward}.
By formulating each operator into a differentiable function, the gradient of $O_h$ back-propagates to $O_l$, $I_l$, and $I_h$ through the computation graph, which enables both the joint training of FCNs and guided filtering layer with direct guidance from the high-resolution targets.
As a result, FCNs can learn to generate a more suitable $O_l$ for guided filtering layer to restore $O_h$.

\begin{algorithm}[!t]
\caption{Gradients for Guided Filtering Layer}
\label{algo:backward}
\SetKwInOut{Input}{Input}
\SetKwInOut{Output}{Output}
\Input{Low-resolution image $I_l$\\High-resolution image $I_h$\\Low-resolution output $O_l$\\Derivative for high-resolution output $\partial O_h$}
\Output{Gradients for all the inputs}
$\begin{aligned}[t]
&\partial b_l=\partial O_h\cdot \nabla_{b_l}f_{\uparrow}\\
&\partial A_l=\partial O_h*G_h\cdot \nabla_{A_l}f_{\uparrow}-\partial b_l*\bar G_l
\end{aligned}$

$\begin{aligned}[t]
&\partial \Sigma_{G_lO_l}=\partial A_l/(\Sigma_{G_l} + \epsilon)\\
&\partial \Sigma_{G_l}=-\partial A_l*\Sigma_{G_lO_l}/(\Sigma_{G_l} + \epsilon)^2
\end{aligned}$

$\begin{aligned}[t]
&\partial \bar O_l=\partial b_l-\partial \Sigma_{G_lO_l}*\bar G_l\\
&\partial O_l=\partial \Sigma_{G_lO_l}\cdot \nabla_{G_l*O_l}f_{\mu}*G_l+\partial \bar O_l\cdot \nabla_{O_l}f_{\mu}
\end{aligned}$

$\begin{aligned}[t]
&\partial \bar G_l=-\partial b_l*A_l-\partial \Sigma_{G_lO_l}*\bar O_l-2\partial \Sigma_{G_l}*\bar G_l
\end{aligned}$

$\begin{aligned}[t]
\partial G_l&=\partial \Sigma_{G_lO_l}\cdot \nabla_{G_l*O_l}f_{\mu}*O_l\\
            &+2\partial \Sigma_{G_l}\cdot \nabla_{G_l*G_l}f_{\mu}*G_l+\partial \bar G_l\cdot \nabla_{G_l}f_{\mu}
\end{aligned}$

$\begin{aligned}[t]
&\partial I_l=\partial G_l\cdot \nabla_{I_l}F\\
&\partial I_h=\partial O_h*A_h\cdot \nabla_{I_h}F
\end{aligned}$
\end{algorithm}

\begin{figure*}[!t]
	\centering
	\includegraphics[width=\linewidth]{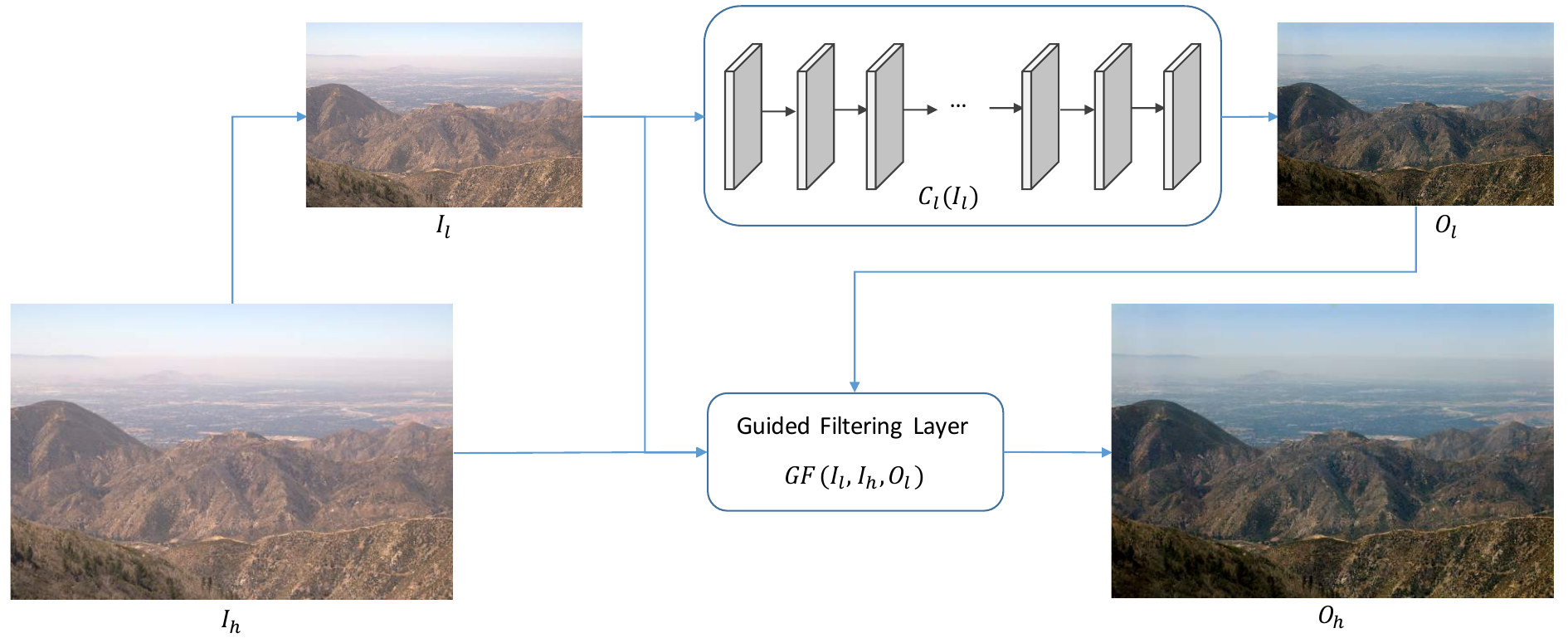}

	\caption{\textbf{Framework Overview of Deep Guided Filtering Network.} Given the input image $I_h$, we first downsample it to obtain $I_l$. The corresponding low-resolution output $O_l$ is then generated by the FCN $C_l(I_l)$. Finally, $I_l$, $I_h$ and $O_l$ are fed into guided filtering layer $GF(I_l,I_h,O_l)$ to generate the high-resolution output $O_h$.}
	\label{fig:framework}
\end{figure*}

\begin{figure}[t]
	\centering
	\includegraphics[width=\linewidth]{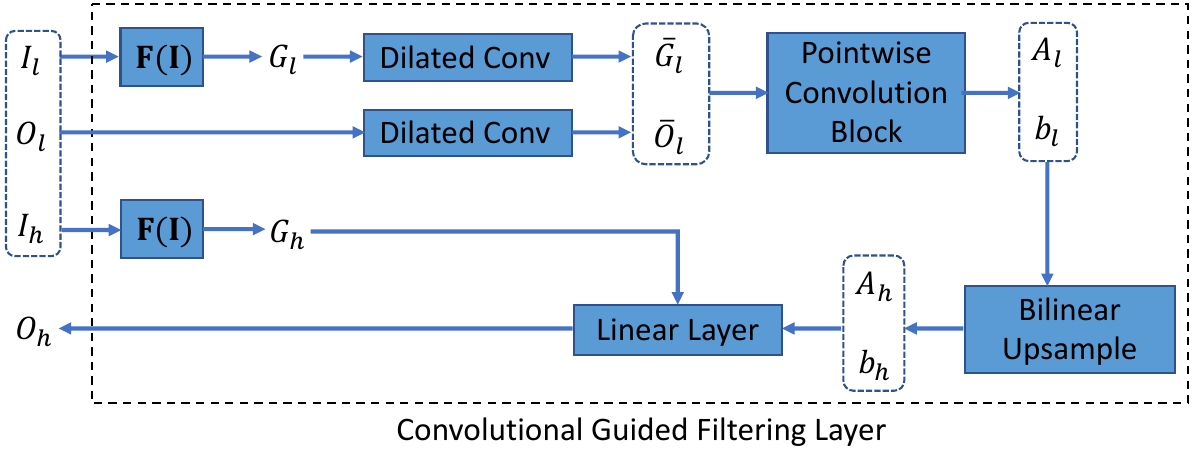}

	\caption{\textbf{Computation Graph of Convolutional Guided Filtering Layer.} Dilated convolutions and a pointwise convolution block are introduced to replace mean filter and local linear model. With learnable parameters, such a layer can adaptively fit a specific task through end-to-end training.}
	\label{fig:conv_guided_filter}
\end{figure}

\subsection{Learn to Generate Task-Specific Guidance Map}
\label{section:dgf}
In Section~\ref{section:dgf_b}, $I_h$, $I_l$ and $O_h$, $O_l$ are assumed to have the same number of channels.
When the channel sizes are different, a transformation function is required to transform $I_h$ and $I_l$ into a guidance map with the same number of channels as $O_h$ and $O_l$.
Even when the channel sizes are the same, a guidance map better than $I_h$ and $I_l$ is necessary for higher performance.
Existing methods usually manually design the transformation function for different tasks, requiring lots of efforts and attempts.
On the contrary, since the proposed guided filtering layer is fully differentiable, we can automatically learn a transformation function to generate more suitable, task-specific guidance maps by end-to-end training.

As shown in Figure~\ref{fig:guided_filter}, the transformation function $F(I)$ transforms $I_h$ and $I_l$ into task-specific guidance maps $G_h$ and $G_l$.
$F(I)$ is a FCN block composed of two convolution layers, between which are an adaptive normalization layer~\cite{Chen/iccv2017} and a leaky ReLU layer.
The kernel size of both convolution layers is set to be $1\times 1$, and the channel size of the first convolution layer is set to be 16 by default.

\subsection{Convolutional Guided Filtering Layer}
\label{section:dgf_conv}

Except $F(I)$, the proposed guided filtering layer is a parameter-free block, which behaves in the same manner for all different tasks.
However, due to the huge differences between tasks, a single guided filtering layer without learnable parameters cannot perform well in all kinds of scenarios.
To solve the problem, we introduce learnable parameters into guided filtering layer by replacing the non-parametric operations into convolution layers.
As a result, the improved layer, convolutional guided filtering layer, becomes more powerful for processing various applications, which can adaptively fit a specific task through end-to-end training.

The architecture of convolutional guided filtering layer is shown in Figure~\ref{fig:conv_guided_filter}.
Compared to that in Figure~\ref{fig:guided_filter}, dilated convolutions are introduced to replace mean filter $f_{\mu}$, and a convolution block composed of pointwise convolutions takes the place of a local linear model. 
As for the hyper-parameters in Section~\ref{section:dgf_b}, $\epsilon$ is removed, and $r$ represents the dilation rates in the dilated convolutions.

\section{Deep Guided Filtering Network}
Based on the proposed guided filtering layer, we present a general framework for pixel-wise image prediction tasks, named Deep Guided Filtering Network (DGF).
By integrating the proposed layer with FCNs following a coarse-to-fine manner, DGF can generate high-resolution, edge-preserving outputs with a much lower computational cost and memory usage.

The architecture of DGF is shown in Figure~\ref{fig:framework}.
First, we downsample the original input image $I_h$ to obtain the low-resolution input $I_l$.
Then, a FCN $C_l(I_l)$ is applied to $I_l$, generating the corresponding low-resolution output $O_l$.
Finally, the high-resolution output $O_h$ is generated by guided filtering layer, taking $I_l$, $I_h$ and $O_l$ as inputs.
The entire network is end-to-end trainable, which could be learned from scratch.

\subsection{Fully Convolutional Network $C_l(I_l)$}
DGF is a general framework for pixel-wise image prediction tasks, which can remarkably reduce the computational complexity and memory usage of the underlying algorithms.
Concretely, given a specific pixel-wise image prediction task, an FCN $C(I)$ can be designed to achieve excellent performance without considering the speed and memory cost.
In order to obtain significant optimization on speed and memory, we can simply drop $C(I)$ into the proposed framework DGF to serve as $C_l(I_l)$ without any other modifications.
Since $C(I)$ processes the input images in low resolution rather than the original resolution, the speed, and memory usage can be largely improved.
Moreover, the performance of our system is also comparable to the previous state-of-the-art one, thanks to the proposed guided filtering layer. This is due to that the proposed guided filtering layer significantly enhances the capabilities of FCNs in the task of joint upsampling.

\subsection{Guided Filtering Layer}
In this paper, there are four variants of DGF in total according to different configurations of the guided filtering layer.
\begin{enumerate}
    \item $\text{DGF}_s$: The original guided filter~\cite{He/2013} is employed as post-processing operation without any training. $C_l(I_l)$ is trained with low-resolution input/output pairs before inserted into $\text{DGF}_s$.
    \item $\text{DGF}_b$: Guided filtering layer in Figure~\ref{fig:guided_filter} is employed in $\text{DGF}_b$. $F(I)$ is an identity function when the inputs and outputs have the same number of channels. When the channel sizes are different, $F(I)$ transforms the inputs into a grey image by averaging along the channel axis. $C_l(I_l)$ and guided filtering layer are jointly trained from scratch under the supervision directly from the high-resolution targets.
    \item $\text{DGF}_b^c$: Compared to $\text{DGF}_b$, the guided filtering layer is replaced by convolutional guided filtering layer in Figure~\ref{fig:conv_guided_filter}.
    \item $\text{DGF}^c$: Compared to $\text{DGF}_b^c$, $F(I)$ proposed in Section~\ref{section:dgf} is introduced, which can learn to generate task-oriented guidance maps without manually design. As a result, $\text{DGF}^c$ is not only end-to-end trainable but can also fit different tasks better by adjusting the trainable convolution weights and the learnable $F(I)$.
\end{enumerate}

\subsection{Objective Function}
DGF is trained end-to-end under the supervision directly from the high-resolution targets.
Concretely, given the high-resolution output $O_h$ and the corresponding target $T_h$, the objective function is defined as $L(O_h, T_h)$.
The concrete formulation varies with different tasks.
Usually, the objective function for training $C(I)$ can be directly employed to train DGF without any adjustment.

\section{Experiments: Image Processing Tasks}
\label{section:experiment_ip}
To show the effectiveness of our method, we employ DGF to clone five widely-used image processing operators.
Concretely, the ground truth images are first generated by applying $L_0$ smoothing operator~\cite{Xu/siggraph2011}, detail manipulation operator~\cite{Farbman/tog2008}, style transfer operator~\cite{Aubry/siggraph2014}, non-local dehazing operator~\cite{Berman/cvpr2016}, and image retouching operator~\cite{fivek} to the input images.
Then, the input/ground-truth pairs are used to train DGF in a supervised way to clone the corresponding image processing operator.

\subsection{Details of Five Image Processing Operators}
\label{section:ops}
\subsubsection{$L_0$ Smoothing}
$L_0$ smoothing~\cite{Xu/siggraph2011} is effective for sharpening major edges while eliminating minor edges by the use of $L_0$ gradient minimization.
To generate the ground truth images, we use the official implementation with the default parameters\footnote{\url{http://www.cse.cuhk.edu.hk/~leojia/projects/L0smoothing}}.

\subsubsection{Detail Manipulation}
Multi-scale detail manipulation~\cite{Farbman/tog2008} enhances an image by boosting features at multiple scales.
Concretely, a three-level decomposition (coarse base level $b$ and two detail levels $d^1$, $d^2$) of the \textbf{CIELAB} lightness channel is first constructed given the input image.
The resulting image is then obtained by a non-linear combination of $b$, $d^1$ and $d^2$.
To generate the ground truth images, we first generate coarse-scale, medium-scale, and fine-scale images with the official implementation and the default parameters\footnote{\url{http://www.cs.huji.ac.il/~danix/epd}}.
The final output is then yielded by averaging the three images.

\subsubsection{Style Transfer}
Photographic style transfer~\cite{Aubry/siggraph2014} aims at transferring the photographic style of a reference image to the input image.
To generate the ground truth images, we employ the official implementation with the default setting and the default reference image\footnote{\url{http://www.di.ens.fr/~aubry/code/matlab_fast_llf_and_style_transfer.zip}}.
The generated outputs are grey images, which are transformed into \textbf{RGB} images as the ground truth.

\subsubsection{Non-local Dehazing}
Non-local dehazing~\cite{Berman/cvpr2016} employs a non-local prior to remove the effects of atmospheric absorption and scattering in the input image.
We use the official implementation with default parameters to generate ground truth images\footnote{\url{https://github.com/danaberman/non-local-dehazing}}.

\subsubsection{Image Retouching}
Image retouching aims at automatically improving the aesthetic quality of the input image by global tonal adjustment.
Human experts are employed to generate the ground truth.

\begin{table}[!t]
\caption{\textbf{The architecture of $C_l(I_l)$ and $F(I)$ for cloning image processing operators.} The negative slope of leaky ReLU is set to $0.2$.}
\label{table:can}
\begin{center}
\begin{tabular}{l|cccccccc|cc}
\hline
&\multicolumn{8}{c}{$C_l(I_l)$}&\multicolumn{2}{|c}{$F(I)$}\\
\hline
Layer & 1 & 2 & 3 & 4 & 5 & 6 & 7 & 8 & 1 & 2\\
\hline
Kernel & 3 & 3 & 3 & 3 & 3 & 3 & 3 & 1 & 3 & 1\\
\#Channels & 24 & 24 & 24 & 24 & 24 & 24 & 24 & 3 & 16 & 3\\
Dilation& 1 & 1 & 2 & 4 & 8 & 16 & 1 & 1 & 1 & 1\\
Bias & \xmark & \xmark & \xmark & \xmark & \xmark & \xmark & \xmark & \checkmark & \xmark & \checkmark\\
AdaptNorm & \checkmark & \checkmark & \checkmark & \checkmark & \checkmark & \checkmark & \checkmark & \xmark & \checkmark & \xmark\\
Leaky ReLU & \checkmark & \checkmark & \checkmark & \checkmark & \checkmark & \checkmark & \checkmark & \xmark & \checkmark & \xmark\\
\hline
\end{tabular}
\end{center}
\end{table}

\begin{figure*}[!t]
	\centering
	\includegraphics[width=\linewidth]{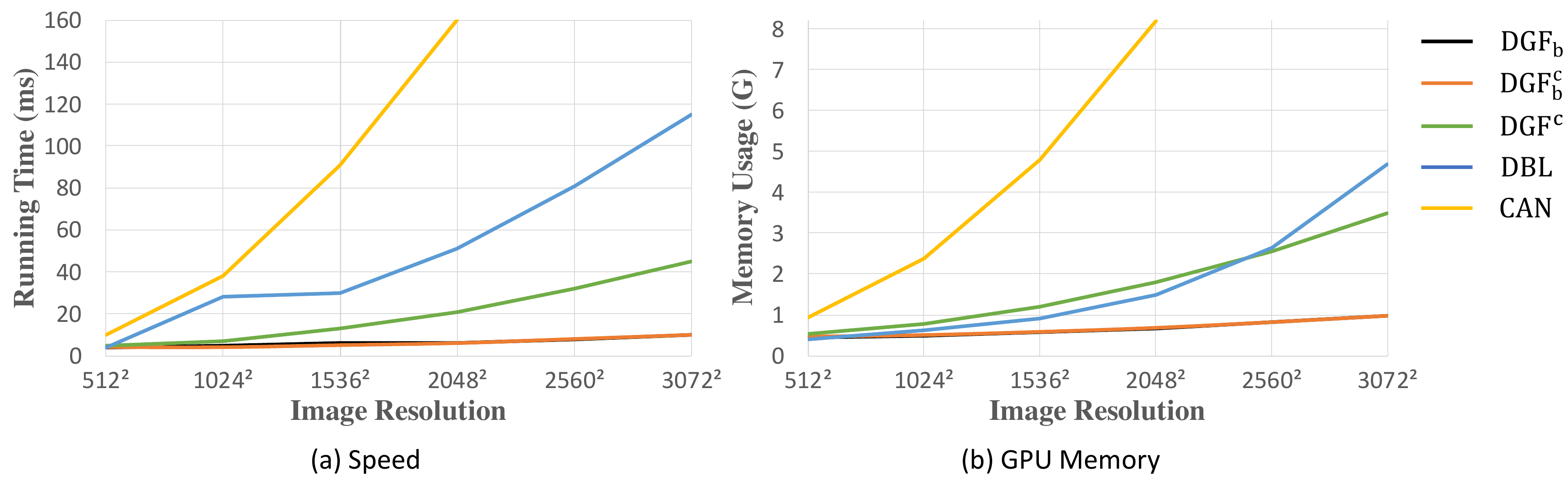}

	\caption{\textbf{Speed and Memory Usage Comparison on GPU Devices.}}
	\label{fig:running_time_and_memory_usage}
\end{figure*}

\subsection{Details of DGF}
We employ Context Aggregation Network (CAN)~\cite{Chen/iccv2017} as $C_l(I_l)$ for all the five image processing operators.
The detailed architectures of $C_l(I_l)$ and $F(I)$ are shown in Table~\ref{table:can}.
AdaptNorm represents adaptive normalization proposed by Chen~\etal~\cite{Chen/iccv2017}.
Leaky ReLU is employed as the nonlinearity, of which the negative slope is set to $0.2$.
As for the objective function, we use $L_2$ loss by following the convention of previous works~\cite{Chen/iccv2017,gharbi2017deep}.

\subsection{Experimental Setup}
Our experiments are taken on MIT-Adobe FiveK Dataset~\cite{fivek}, which contains 2,500/2,500 high-resolution photographs as the training/testing images.
In the dataset, each photograph contains five annotations from five experts, which can be used as the ground truth for image retouching.
Instead of all five annotations, we employ the annotation from expert A as the ground truth.
As for the other four image processing operators, the ground truth images are generated following the instructions in Section~\ref{section:ops}.

As for training, we first train the network for 150 epochs, with the input/target images resized to 512\textbf{s}\footnote{$x$\textbf{s} means the short side of an image is resized to $x$ without changing the aspect ratio.}.
To improve the generalization ability, we further train the network for 30 epochs, with training data randomly resized to a specific resolution between 512\textbf{s} and 1672\text{s}.
As for $I_l$, the spatial resolution is 64\textbf{s} regardless of the resolution of $I_h$.
Adam is employed as the optimizer, with learning rate set to $0.0001$ and batch size set to $1$.

Our primary baseline is Deep Bilateral Learning (DBL)~\cite{gharbi2017deep}, which shares a similar architecture to ours and achieves a good trade-off between quality and speed.
Another strong baseline is CAN~\cite{Chen/iccv2017}, which achieves state-of-the-art performance while runs reasonably fast.
To ensure a fair comparison, we train the models using the official implementations and training procedures for both methods.

\subsection{Experimental Results}

\begin{table*}[ht]
\caption{\textbf{Quantitative Comparison on Image Processing Tasks.} The 1st, 2nd and 3rd methods are highlighted as red, green, and blue.
}
\label{table:mse_psnr_ssim}
\centering
\begin{tabular}{l|r|r|r|r|r|r|r|r|r|r|r|r|r|r|r}
\hline
\multirow{2}{*}{Method} & \multicolumn{3}{c}{$L_0$ Smoothing~\cite{Xu/siggraph2011}} & \multicolumn{3}{|c}{Detail Manipulation~\cite{Farbman/tog2008}} & \multicolumn{3}{|c}{Style Transfer~\cite{Aubry/siggraph2014}} & \multicolumn{3}{|c}{Non-local Dehazing~\cite{Berman/cvpr2016}} & \multicolumn{3}{|c}{Image Retouching~\cite{fivek}} \\
\cline{2-16}
& MSE & PSNR & SSIM & MSE & PSNR & SSIM & MSE & PSNR & SSIM & MSE & PSNR & SSIM & MSE & PSNR & SSIM\\
\hline
Input   & 73 & 29.61 & 0.796 & 443 & 22.12 & 0.789 & 3534 & 13.28 & 0.521 & 2081 & 16.95 & 0.684 & 1507 & 18.44 & 0.727\\
\hline
CAN~\cite{Chen/iccv2017}     & \textcolor{red}{27} & \textcolor{red}{35.05} & \textcolor{red}{0.970} & \textcolor{red}{9} & \textcolor{red}{38.97} & \textcolor{red}{0.986} & 519 & 21.31 & \textcolor{green}{0.870} & 355 & 24.47 & 0.862 &  964 & 20.43 & 0.744\\
\hline
DBL~\cite{gharbi2017deep}     & 39 & 32.35 & 0.896 & 75 & 29.84 & 0.924 & 354 & 23.32 & 0.834 & 502 & 23.27 & 0.852 & 1056 & 20.21 & 0.748\\
\hline
DJF~\cite{li2016deep}     & 90 & 29.40 & \textcolor{green}{0.937} & 100 & 28.99 & \textcolor{blue}{0.927} & 383 & 22.73 & \textcolor{blue}{0.856} & 649 & 21.04 & 0.724 & 1216 & 18.89 & 0.702\\
\hline
\hline
$\text{DGF}_\text{s}$ & 35 & 32.93 & 0.912 & 92 & 29.12 & 0.905 & 333 & 23.22 & 0.735 & 351 & 24.53 & 0.871 & \textcolor{green}{872} & 20.81 & 0.757\\
\hline
$\text{DGF}_\text{b}$ & 33 & 33.20 & 0.911 & 77 & 29.95 & 0.905 & \textcolor{blue}{318} & \textcolor{blue}{23.42} & 0.738 & \textcolor{blue}{323} & \textcolor{blue}{25.53} & \textcolor{blue}{0.892} & \textcolor{blue}{875} & \textcolor{blue}{20.94} & \textcolor{green}{0.762}\\
\hline
$\text{DGF}_\text{b}^\text{c}$ & \textcolor{blue}{32} & \textcolor{blue}{33.39} & 0.917 & \textcolor{blue}{69} & \textcolor{blue}{30.48} & 0.916 & \textcolor{green}{305} & \textcolor{green}{23.61} & 0.751 & \textcolor{red}{293} & \textcolor{green}{25.76} & \textcolor{green}{0.896} & \textcolor{red}{855} & \textcolor{green}{21.04} & \textcolor{blue}{0.760}\\
\hline
$\text{DGF}^\text{c}$ & \textcolor{green}{30} & \textcolor{green}{33.69} & \textcolor{blue}{0.923} & \textcolor{green}{48} & \textcolor{green}{32.10} & \textcolor{green}{0.940} & \textcolor{red}{181} & \textcolor{red}{26.17} & \textcolor{red}{0.880} & \textcolor{green}{296} & \textcolor{red}{25.85} & \textcolor{red}{0.902} & \textcolor{red}{855} & \textcolor{red}{21.09} & \textcolor{red}{0.767}\\
\hline
\end{tabular}
\end{table*}

\begin{table*}[ht]
\caption{\textbf{Upper Bound for Our Method on Image Processing Tasks.}}
\label{table:upper}
\centering
\begin{tabular}{l|r|r|r|r|r|r|r|r|r|r|r|r|r|r|r}
\hline
\multirow{2}{*}{Method} & \multicolumn{3}{c}{$L_0$ Smoothing~\cite{Xu/siggraph2011}} & \multicolumn{3}{|c}{Detail Manipulation~\cite{Farbman/tog2008}} & \multicolumn{3}{|c}{Style Transfer~\cite{Aubry/siggraph2014}} & \multicolumn{3}{|c}{Non-local Dehazing~\cite{Berman/cvpr2016}} & \multicolumn{3}{|c}{Image Retouching~\cite{fivek}} \\
\cline{2-16}
& MSE & PSNR & SSIM & MSE & PSNR & SSIM & MSE & PSNR & SSIM & MSE & PSNR & SSIM & MSE & PSNR & SSIM\\
\hline
$\text{DGF}_\text{b}$ & 27 & 34.52 & 0.923 & 93 & 29.23 & 0.904 & 261 & 24.28 & 0.752 & 49 & 33.00 & 0.956 & 100 & 30.82 & 0.859	\\
\hline
$\text{DGF}_\text{b}^\text{c}$ & 25 & 34.93 & 0.925 & 71 & 30.35 & 0.917 & 245 & 24.55 & 0.761 & 34 & 34.35 & 0.964 & 94 & 30.97 & 0.857\\
\hline
$\text{DGF}^\text{c}$ & 23 &	35.27 &	0.930 & 42 & 32.66 & 0.947 & 109 & 28.34 & 0.899 & 28 & 35.19 & 0.968 & 90 & 31.19 & 0.860\\
\hline
\end{tabular}
\end{table*}

\subsubsection{Running Time and Memory Usage}
The running time and memory usage are shown in Figure~\ref{fig:running_time_and_memory_usage}, which are measured on a workstation with Intel E5-2650 2.20GHz CPU and Nvidia Titan X (Pascal) GPU.

On GPU devices, both $\text{DGF}_\text{b}$ and $\text{DGF}_\text{b}^\text{c}$ take less than 10ms to process an image with resolution ranging from $512^2$ to $3072^2$.
$\text{DGF}^\text{c}$ is slightly slower because of the usage of $F(I)$, but it still runs in real-time on images with resolution $3072^2$.
All the three variants of our method run much faster than CAN and DBL among all resolutions.
Specifically, $\text{DGF}_\text{b}$, $\text{DGF}_\text{b}^\text{c}$, and $\text{DGF}^\text{c}$ take 6ms, 6ms, and 21ms respectively for an image in $2048^2$.
CAN takes 160ms for an image in $2048^2$, which are more than $25\times$, $25\times$, and $7\times$ slower than our method.
DBL takes 51ms in the same setting, which is slightly faster than CAN but more than $8\times$ slower than $\text{DGF}_\text{b}$ and $\text{DGF}_\text{b}^\text{c}$.
The advantage of our method in speed is even more significant as the resolution grows.

For $I_h$ with $h \times w \times n_I$ and $O_h$ with $h \times w \times n_O$, the theoretical computational complexities of $\text{DGF}_\text{b}$, $\text{DGF}_\text{b}^\text{c}$, $\text{DGF}^\text{c}$, and $\text{DBL}$ are $\mathcal{O}(n_O\times h \times w)$, $\mathcal{O}(n_O\times h \times w)$, $\mathcal{O} ( (n_I+n_O)\times h\times w )$ and $\mathcal{O} ( n_I\times n_O\times h \times w )$ respectively.

As for memory usage, our method takes less GPU memory space than both baseline methods.
CAN is the most memory inefficient method that takes nearly 10G GPU memory to process an image with resolution $2048^2$.
$\text{DGF}^\text{c}$ takes a similar amount of memory space to that of DBL but grows slower as the resolution increases.
$\text{DGF}_\text{b}$ and $\text{DGF}_\text{b}^\text{c}$ are the most memory efficient methods, which take less than 1G memory even on images with resolution $3072^2$.

\subsubsection{Quantitative and Qualitative Comparison}
The performance of each method is evaluated on the test set of MIT-Adobe FiveK dataset with input/target images resized to 1024\textbf{s}.
MSE, PSNR, and SSIM serve as the evaluation metrics.

As shown in Table~\ref{table:mse_psnr_ssim}, our method achieves the state-of-the-art performance in style transfer, non-local dehazing, and image retouching; while obtaining comparable results in $L_0$ smoothing and multi-scale detail manipulation.
Concretely, $\text{DGF}^\text{c}$ achieves 26.17 dB in PSNR for style transfer, which improves over CAN and DBL by 4.86 dB and 2.85 dB respectively.
Compared to DBL, our method outperforms it across all the five tasks in all three metrics by a large margin.

\begin{figure*}[!t]
	\captionsetup[subfigure]{labelformat=empty}
	\centering
	\subfloat[Non-local Dehazing~\cite{Berman/cvpr2016}]{\includegraphics[width=0.5\linewidth]{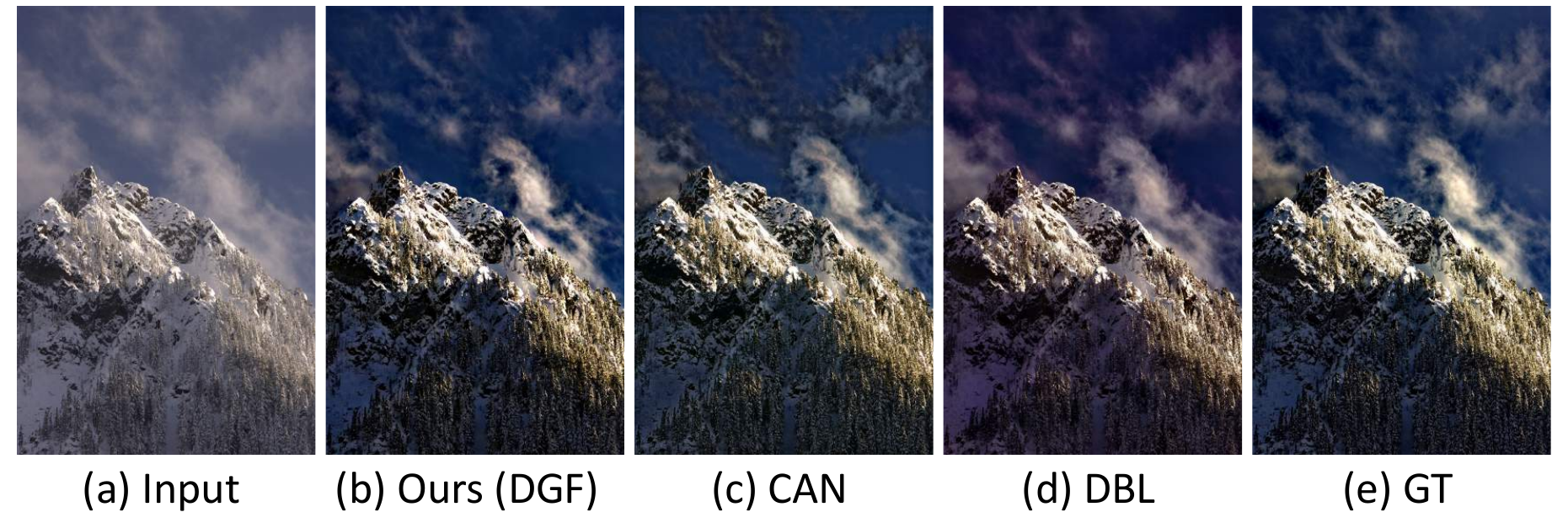}}
	\subfloat[Style Transfer~\cite{Aubry/siggraph2014}]{\includegraphics[width=0.5\linewidth]{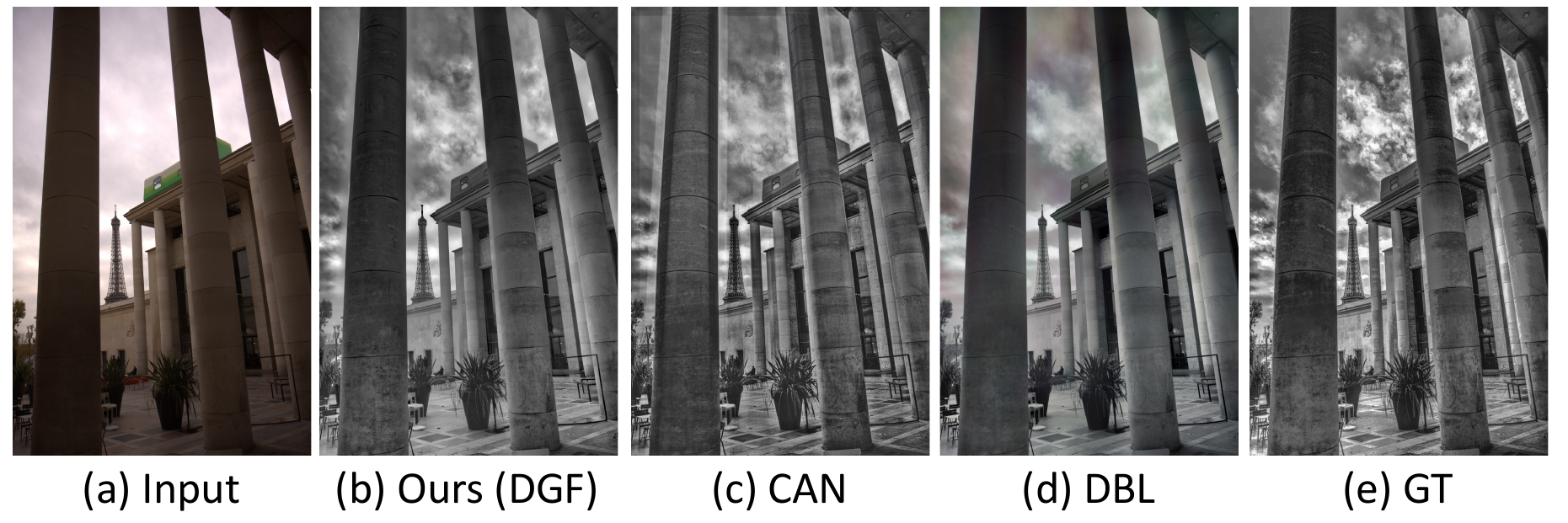}}\\
	\subfloat[Multi-scale Detail Manipulation~\cite{Farbman/tog2008}]{\includegraphics[width=0.5\linewidth]{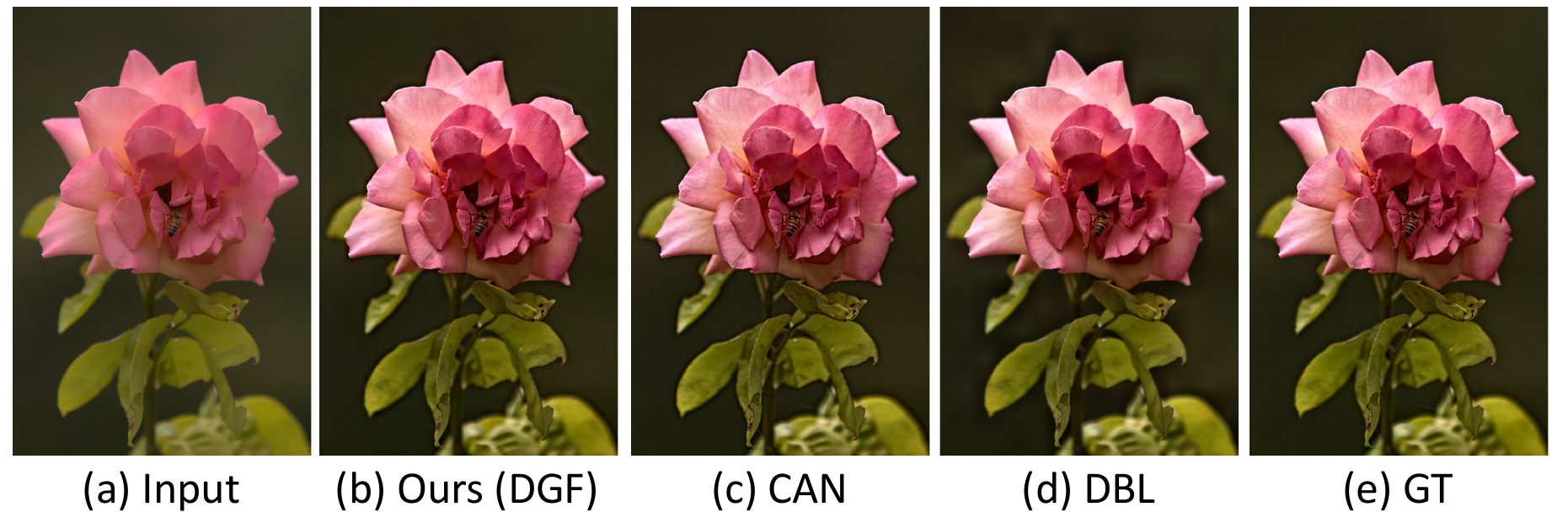}}
	\subfloat[$L_0$ Smoothing~\cite{Xu/siggraph2011}]{\includegraphics[width=0.5\linewidth]{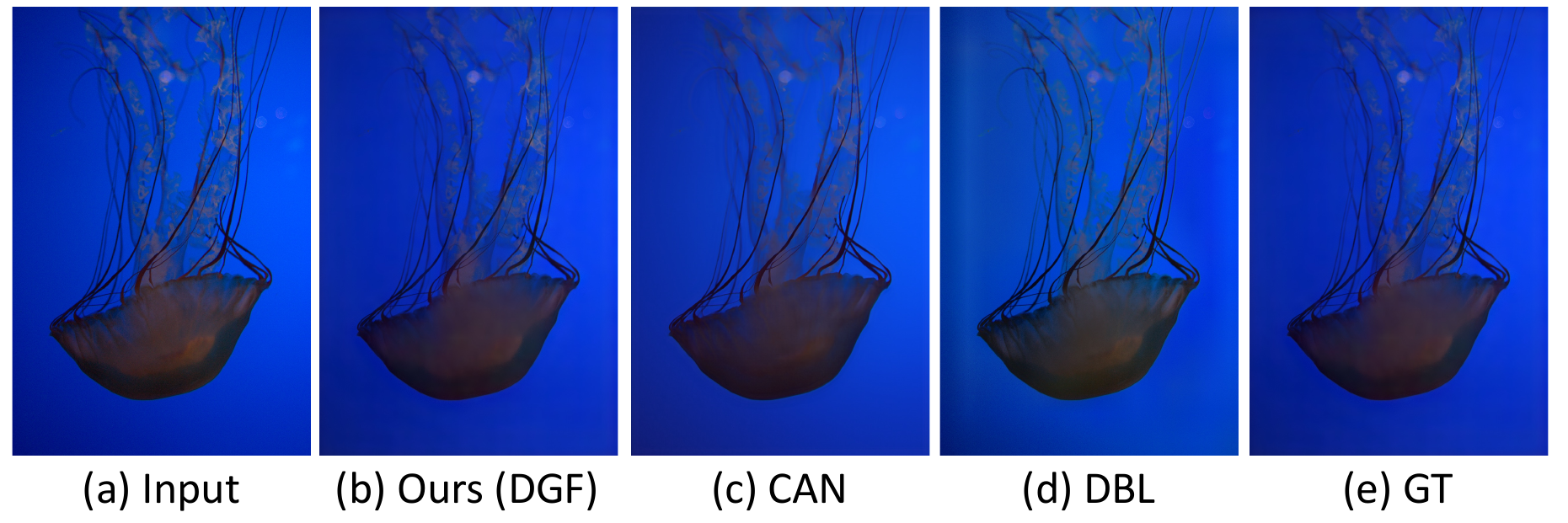}}\\
	\caption{\textbf{Qualitative Results in Image Processing.} Our method $\text{DGF}^\text{c}$ is more visually appealing than other approaches. Best viewed in color.}
	\label{fig:qualitive}
\end{figure*}

The qualitative results are shown in Figure~\ref{fig:qualitive}\footnote{More qualitative results are shown in \url{http://wuhuikai.me/DeepGuidedFilterProject/#visual}.}.

\subsubsection{The Role of Guided Filtering Layer}
To show the effect of convolutional guided filtering layer and $F(I)$, we replace $O_l$ with low-resolution ground truth to generate $O_h$.
The obtained result represents the performance upper bound of each DGF variant.
As shown in Table~\ref{table:upper}, by reformulating guided filtering layer into learnable convolution layers, $\text{DGF}_\text{b}^\text{c}$ outperforms $\text{DGF}_\text{b}$ in all five tasks.
By further introducing $F(I)$ into the convolutional guided filtering layer, $\text{DGF}^\text{c}$ achieves the best performance.

Similar results can be observed in Table~\ref{table:mse_psnr_ssim}.
By jointly end-to-end training, $\text{DGF}_\text{b}$ achieves better performance on most tasks than $\text{DGF}_\text{s}$.
Concretely, $\text{DGF}_\text{b}$ improves 1 dB and 0.83 dB (PSNR) for non-local dehazing and detail manipulation.
By reformulating into convolutional guided filtering layer, the performance is further improved by comparing $\text{DGF}_\text{b}$ and $\text{DGF}_\text{b}^\text{c}$.
By adding learnable $F(I)$, we gain significant improvements in several tasks, especially in tasks that are resolution-dependent.
Table~\ref{table:mse_psnr_ssim} shows that $\text{DGF}^\text{c}$ increases PSNR by 2.56 dB and 1.62 dB compared to $\text{DGF}_\text{b}^\text{c}$ for style transfer and detail manipulation.

DJF~\cite{li2016deep} is the state-of-the-art method for joint upsampling.
To verify the effectiveness of our method, we replace guided filtering layer in DGF with DJF.
Results in Table~\ref{table:mse_psnr_ssim} show that our method outperforms DJF in all tasks.
Besides, our method also runs much faster than DJF, which takes $9\times$ less time than DJF on images with resolution $1024^2$ (5ms v.s. 46ms).

\begin{figure}[!t]
	\centering
	\includegraphics[width=\linewidth]{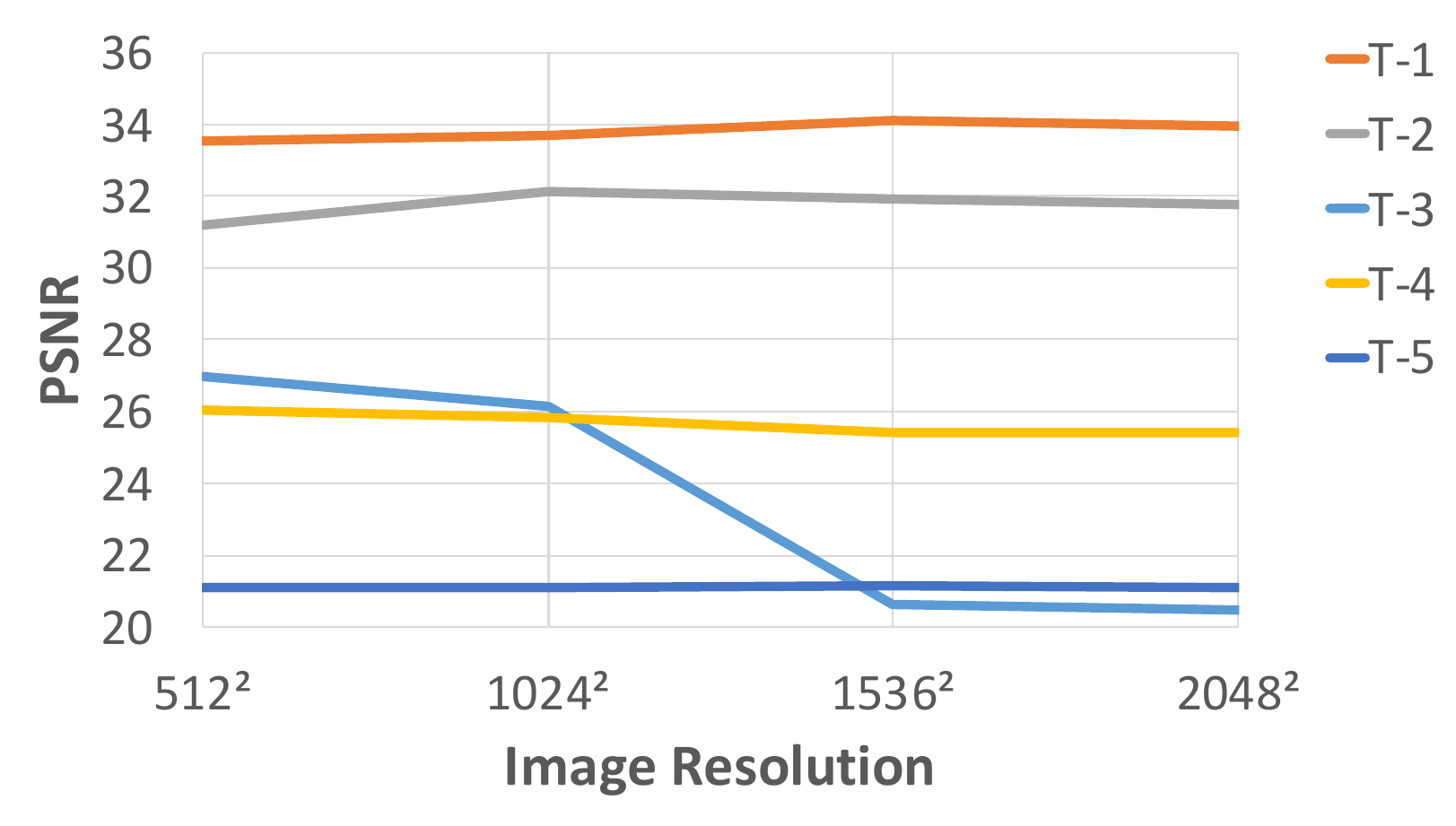}

	\caption{\textbf{Cross Resolution Generalization.} T-$x$ represents the $x$-th image processing task in Table~\ref{table:mse_psnr_ssim}.}
	\label{fig:generalization}
\end{figure}

\subsubsection{Cross Resolution Generalization}
In the main experiment, our method is evaluated on 1024\textbf{s} images.
To show the generalization ability of DGF for processing images in different resolutions, the pre-trained DGF is directed employed on images in 512\textbf{s}, 1024\textbf{s}, 1536\textbf{s}, and 2048\textbf{s} without finetuning.
As shown in Figure~\ref{fig:generalization}, our method performs equally well across different resolutions on all tasks except style transfer.
The reason is that style transfer is highly resolution-dependent.
Concretely, given a reference image with a fixed resolution, the styles of the outputs are different for input images with different resolutions. 

\subsubsection{Ablation Study} 
A series of experiments are taken in this section to validate the effect of each hyper-parameter in the proposed guided filtering layer.

\begin{figure*}[!t]
	\centering
	\subfloat[Radius $r$]{\includegraphics[width=0.33\linewidth]{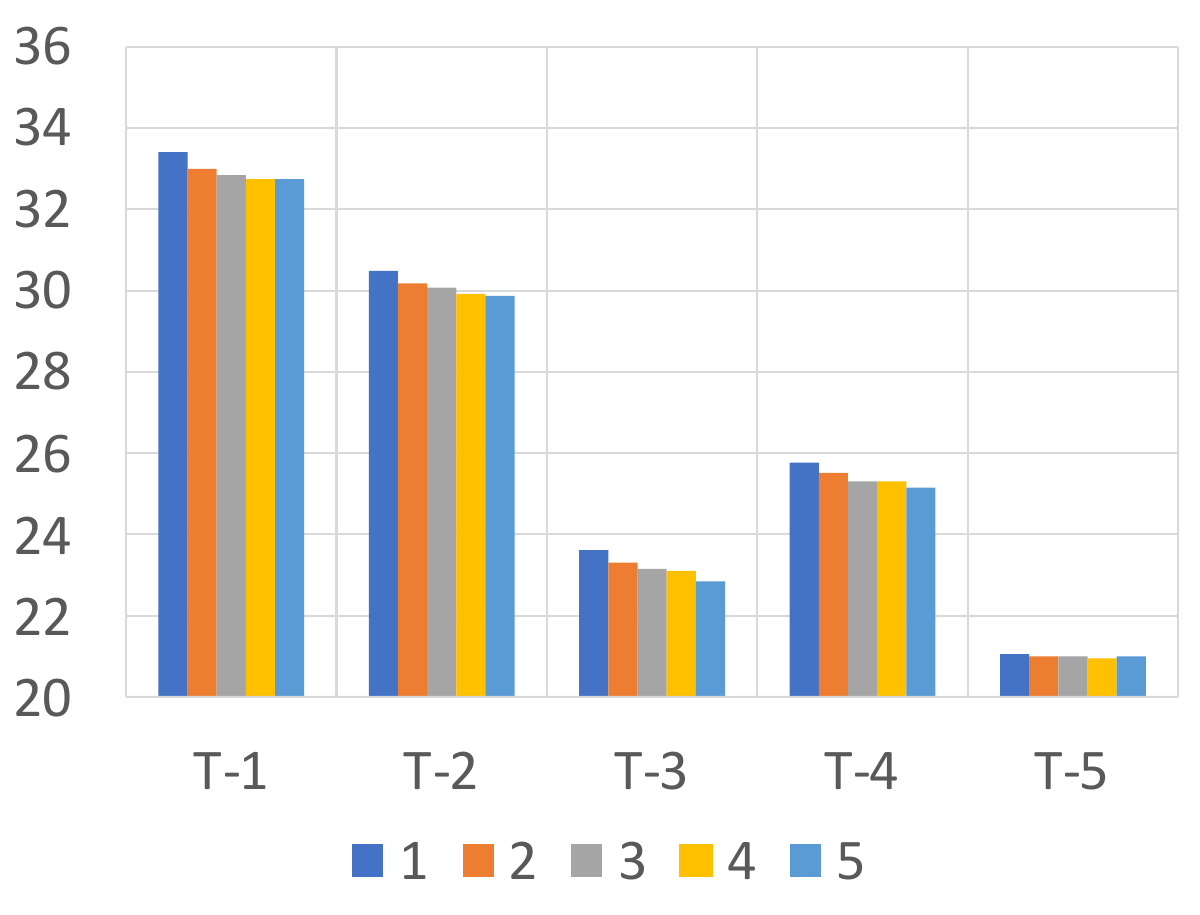}\label{fig:ablation_r}}
	\subfloat[Resolution of $I_l$]{\includegraphics[width=0.33\linewidth]{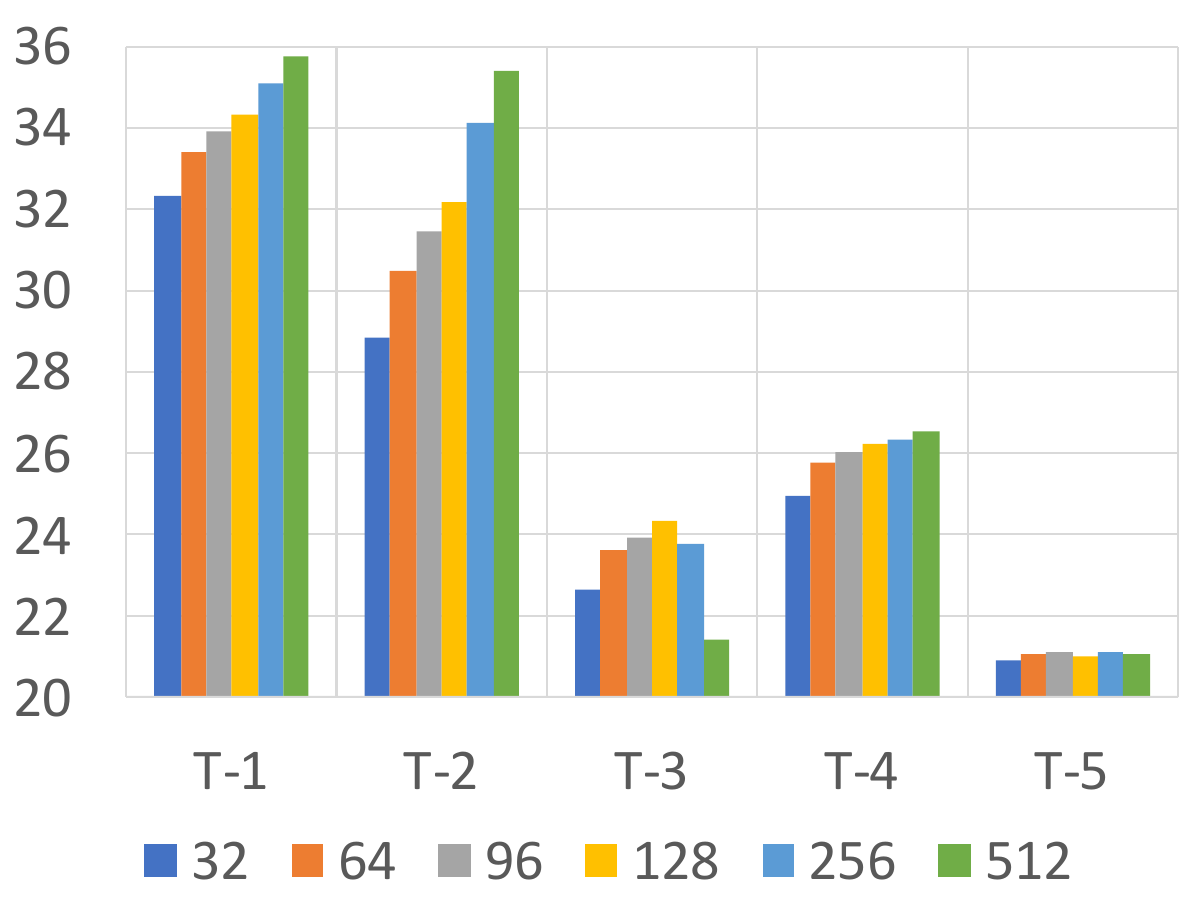}\label{fig:ablation_lr}}
	\subfloat[Dilation Rate of $F(I)$]{\includegraphics[width=0.33\linewidth]{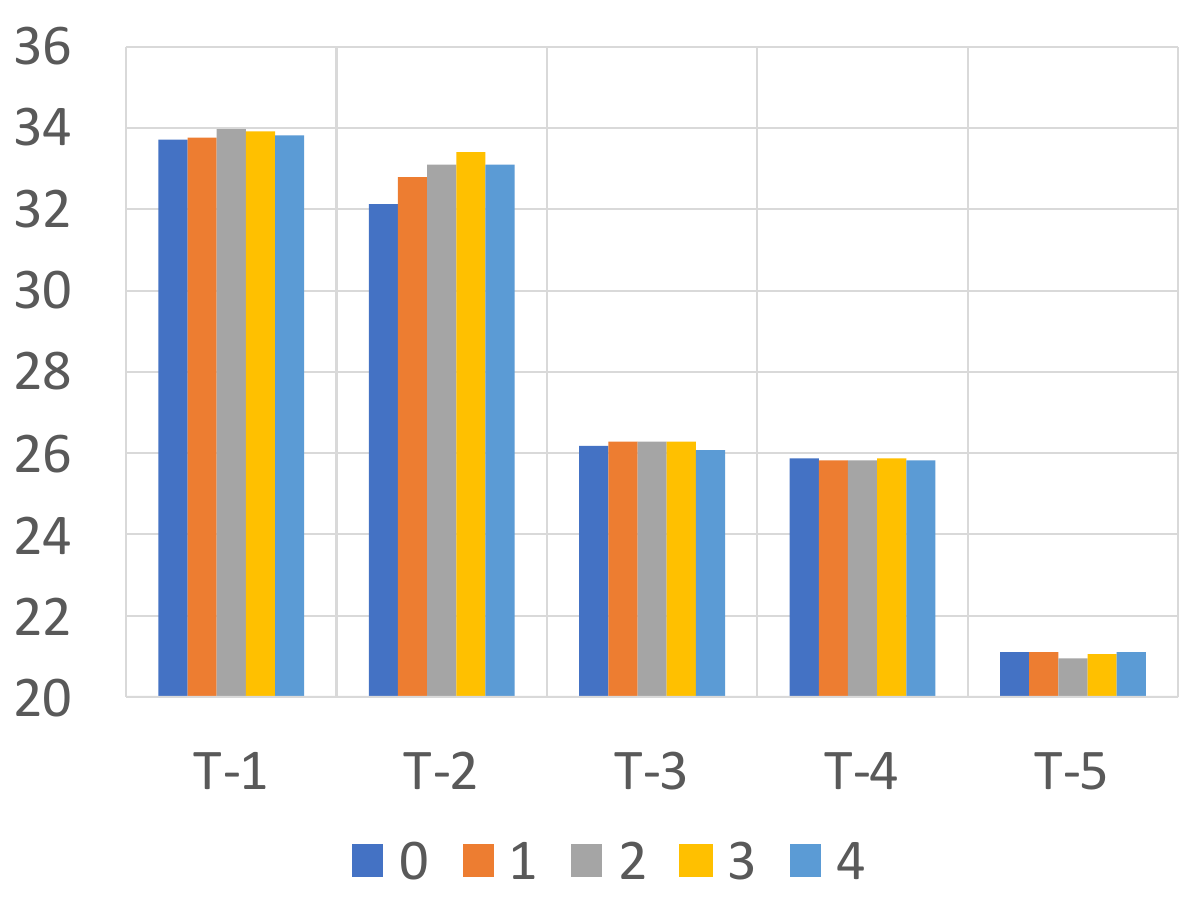}\label{fig:ablation_dilation}}

	\caption{\textbf{Ablation Study.} The performance (PSNR) with different (a) radius $r$ of guided filtering layer, (b) resolution of $I_l$, and (c) dilation rate of $F(I)$. T-$x$ represents the $x$-th image processing task in Table~\ref{table:mse_psnr_ssim}. Best viewed in color.}
	\label{fig:ablation}
\end{figure*}

The role of radius $r$ is shown in Figure~\ref{fig:ablation_r}.
The performance drops quickly as $r$ grows, and the default setting ($r = 1$) obtains the best PSNR score.

The effect of the resolution of $I_l$ is shown in Figure~\ref{fig:ablation_lr}.
For $L_0$ smoothing, multi-scale detail manipulation, and non-local dehazing, the performance grows as the resolution of $I_l$ increases.
For style transfer and image retouching, higher resolution is not always better.
The corresponding running time and memory usage are shown in Table~\ref{table:speed_memory_lr}.
When the resolution of $I_l$ is 128 or 256, our method can not only achieve an excellent performance but also run very fast.

The function of $F(I)$ is also explored by varying the dilation rate.
Figure~\ref{fig:ablation_dilation} shows that increasing the dilation rate can improve the performance to a degree.

\begin{table}
\caption{\textbf{Speed and Memory Usage in Different Resolutions of $I_l$.}
}
\label{table:speed_memory_lr}
\centering
\begin{tabular}{l|c|c|c|c|c|c}
\hline
Resolution & 32 & 64 & 96 & 128 & 256 & 512 \\
\hline
Running Time (ms) & 6 & 6 & 6 & 7 & 7 & 19 \\
\hline
Memory Usage (G) & 0.67 & 0.68 & 0.70 & 0.73 & 0.87 & 1.41 \\
\hline
\end{tabular}
\end{table}

\begin{figure*}[!t]
	\captionsetup[subfigure]{labelformat=empty}
	\centering
	\subfloat[Saliency Object Detection~\cite{Liu/cvpr2007}]{\includegraphics[width=0.5\linewidth]{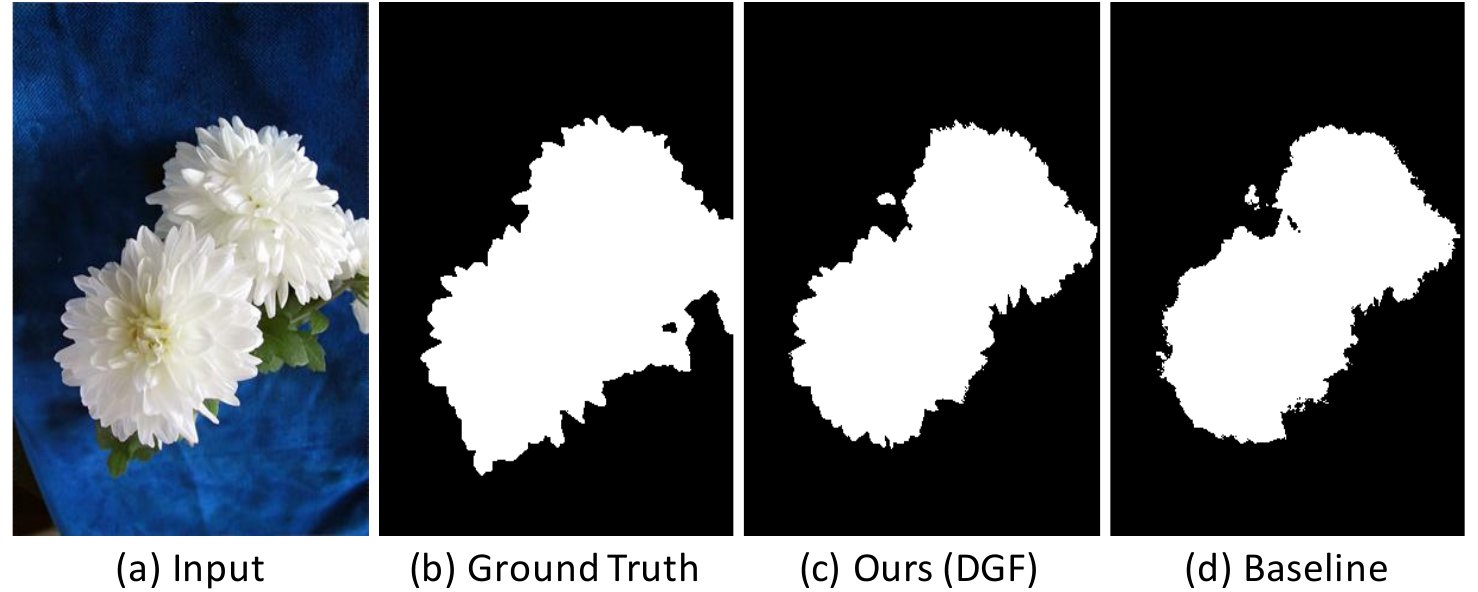}}
	\subfloat[Semantic Segmentation~\cite{He/cvpr2004}]{\includegraphics[width=0.5\linewidth]{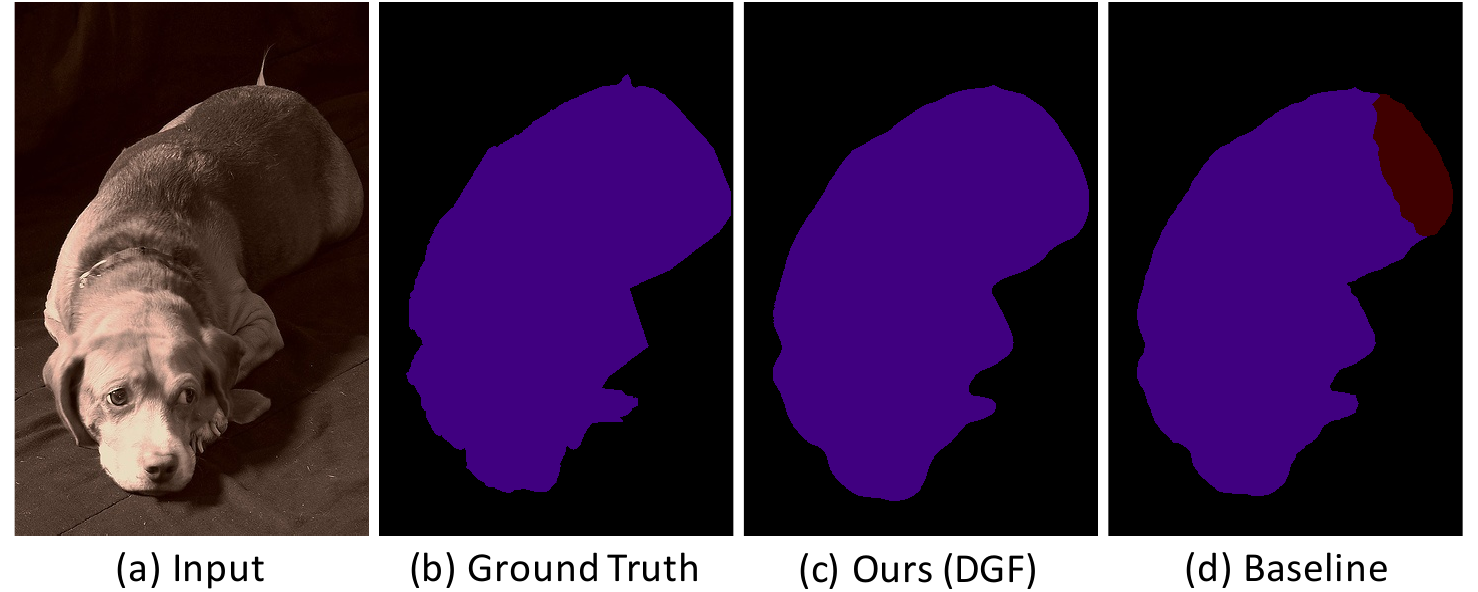}}\\
	\caption{\textbf{Qualitative Results in Computer Vision.} Best viewed in color.}
	\label{fig:qualitive_cv}
\end{figure*}

\section{Experiments: Computer Vision Tasks}
The proposed guided filtering layer can dramatically advance the performance of multiple image processing tasks in accuracy, speed, and memory usage.
Moreover, our method can also be employed to replace the time-consuming conditional random field (CRF) in many computer vision applications.
To evaluate the effectiveness of our method, we take an experiment on three computer vision tasks ranging from low-level vision to high level-vision, namely depth estimation~\cite{Saxena/nips2005}, saliency object detection~\cite{Liu/cvpr2007}, and semantic segmentation~\cite{He/cvpr2004}.

\subsection{Details of Three Computer Vision Tasks}
\subsubsection{Depth Estimation}
Depth estimation is proposed by Saxena~\etal~\cite{Saxena/nips2005}, which aims at predicting the depth at each pixel of an image with monocular cues.
For this task, KITTI~\cite{geiger2012we} is the most widely used dataset, which contains 42,382 rectified stereo pairs from 61 scenes.
In this paper, 29,000/1,159 images from the official training set are used for training and evaluation, which covers 33 scenes.
The remaining 28 scenes of the official training set contain 200 high-quality disparity images, which are used for testing in this paper.

\subsubsection{Saliency Object Detection}
Saliency object detection is used to detect the most salient object in an image, which is formulated as an image segmentation problem~\cite{Liu/cvpr2007}.
MSRA-B~\cite{jiang2013salient} and the official training/validation/test split~\cite{jiang2013salient} is used in our experiment.

\subsubsection{Semantic Segmentation}
Semantic segmentation aims at assigning each pixel of an image to one of the pre-defined labels~\cite{He/cvpr2004}.
To evaluate our method, PASCAL VOC 2012 benchmark~\cite{pascal-voc-2012} is used in this paper, which involves 20 foreground object classes and one background class.
The original dataset contains 1,464, 1,449, and 1,456 pixel-wise labeled images for training, validation, and testing, respectively.
The training set is further augmented by extra annotations~\cite{hariharan2011semantic}, resulting in 10,582 images.
We use the 10,582 augmented images for training and the 1,449 validation images for testing.

\subsection{Details of DGF}
When applying DGF to computer vision tasks, the high-resolution input image is directly processed by $C_l(I_l)$ without downsampling, generating the low-resolution output $O_l$.
As for the architecture of $C_l(I_l)$, MonoDepth\footnote{\url{https://github.com/mrharicot/monodepth}}~\cite{godard2017unsupervised}, DSS\footnote{\url{https://github.com/wuhuikai/DeepGuidedFilter}}~\cite{hou2017deeply}, DeepLab-V2\footnote{\url{https://github.com/isht7/pytorch-deeplab-resnet}}~\cite{chen2016deeplab} are employed for depth estimation, saliency detection, and semantic segmentation respectively.
The corresponding training and testing procedures and loss functions are also used to train our network.
As for the hyper-parameters of guided filtering layer, $r$ and $\epsilon$ are determined by grid search on the validation set, as shown in Table~\ref{table:parameter}.
Notably, a second guided filtering layer is applied in the saliency detection task to achieve better performance.

\subsection{Main Results}
The performances of our method and baseline methods are shown in Table~\ref{table:vision}.
For depth estimation, $\text{DGF}_\text{s}$ obtain 0.177 improvements in rms over the baseline.
By end-to-end training and adding the learnable guidance map, we achieve the best performance (5.887) in rms.
Similar results are obtained in saliency detection and semantic segmentation.
$F_\beta$ increases from 90.61\% to 91.29\% by applying the guided filtering layer to saliency detection.
By replacing $\text{DGF}_\text{s}$ with DGF, $F_\beta$ further improves to 91.75\%.
For segmentation, DGF obtains 73.58\% in mean IOU, which has an improvement of 1.79\% compared to the baseline method.

We also compare our method with DenseCRF~\cite{krahenbuhl2011efficient}, which is commonly used in saliency detection and semantic segmentation.
Experiments show that our method is comparable to DenseCRF in saliency detection, and obtains better performance in semantic segmentation. 
Besides, the proposed layer performs at least $10\times$ faster than DenseCRF. Averagely, our approach takes $34$ms to process a $512^2$ image, while DenseCRF takes $432$ms.

Figure~\ref{fig:qualitive_cv} shows the visual results of our method and baselines. The results obtained by our approach are better in preserving edges and details\footnote{More qualitative results are shown in \url{http://wuhuikai.me/DeepGuidedFilterProject/#visual}.}.

\begin{table}
\caption{\textbf{Hyper-Parameters of Guided Filtering Layer.}}
\label{table:parameter}
\begin{center}
\begin{tabular}{l|c|c|c|c}
\hline
&\multicolumn{2}{c}{1st guided filtering layer}&\multicolumn{2}{|c}{2nd guided filtering layer}\\
\hline
Hyper-Params & $r$ & $\epsilon$ & $r$ & $\epsilon$\\
\hline
Depth & 4 & 1e-2 & - & -\\
Saliency & 8 & 1e-2 & 8 & 1e-2\\
Segmentation & 4 & 1e-2 & - & -\\
\hline
\end{tabular}
\end{center}
\end{table}

\begin{table}
\caption{\textbf{Quantitative Comparison on Computer Vision Tasks.}}
\label{table:vision}
\centering
\begin{tabular}{l|c|c|c|c}
\hline
\multirow{2}{*}{Method}&\multicolumn{2}{|c|}{Depth Estimation}&Saliency Detection&Segmentation\\
\cline{2-5}
&rms&log10&$F_\beta$&Mean IOU\\
\hline
\hline
Baseline&6.081&0.216&90.61\%&71.79\%\\
\hline
DenseCRF&-&-&\textbf{91.87\%}&72.69\%\\
\hline
\hline
$\text{DGF}_\text{s}$&5.904&0.211&91.29\%&71.72\%\\
\hline
DGF&\textbf{5.887}&\textbf{0.209}&91.75\%&\textbf{73.58\%}\\
\hline
\end{tabular}
\end{table}

\section{Conclusion}
We present a novel building block for FCN, namely guided filtering layer, which aims at enhancing the ability of FCNs for joint upsampling.
By formulating the guided filter into a fully differentiable module with learnable convolutional kernels, FCN-based pixel-wise image prediction approaches can benefit from end-to-end training and generate high-quality results.
We further extend the proposed layer with a learnable transformation function, which makes it generalize well to different tasks by producing task-specific guidance maps.
We integrate the guided filtering layer with FCNs and evaluate it on five image processing tasks and three computer vision tasks.
Experiments show that the proposed layer could achieve state-of-the-art performance while taking $10\text{-}100\times$ less computational cost.
We also conduct a comprehensive ablation study, which demonstrates the contribution of each component as well as the hyper-parameters.

\section*{Acknowledgement}
This work is funded by the National Key Research and Development Program of China (Grant 2016YFB1001004 and Grant 2016YFB1001005), the National Natural Science Foundation of China (Grant 61673375, Grant 61721004 and Grant 61403383) and the Projects of Chinese Academy of Sciences (Grant QYZDB-SSW-JSC006 and Grant 173211KYS-B20160008). The authors would like to thank Patrick P{\'e}rez and Philip Torr for their helpful suggestions.

\bibliographystyle{IEEEtran}
\bibliography{IEEEabrv,ref}

\end{document}